%% file: iclr2027_conference.tex
\documentclass{article} 
\usepackage{iclr2027_conference,times}
\input{math_commands.tex}

\usepackage{xspace}
\usepackage{algorithm}
\usepackage{algpseudocode}
\usepackage{hyperref}
\usepackage{url}
\usepackage{booktabs}
\usepackage[table]{xcolor}
\usepackage{subcaption}
\usepackage{xltabular}
\usepackage{graphicx}
\usepackage{colortbl}
\definecolor{HeaderTeal}{RGB}{44,91,93}
\definecolor{ScoreMint}{RGB}{176,224,215}
\definecolor{Qwen3Header}{RGB}{44,91,93}
\definecolor{Qwen3Mint}{RGB}{195,231,224}
\definecolor{Qwen25Header}{RGB}{91,72,120}
\definecolor{Qwen25Lavender}{RGB}{224,211,239}
\definecolor{SmolHeadedr}{RGB}{145,82,45}
\definecolor{SmolPeach}{RGB}{247,218,194}
\usepackage{multirow}
\usepackage{longtable}
\usepackage{wrapfig}
\usepackage{booktabs}
\usepackage{tabularx}
\usepackage{makecell}

\title{A Strong Baseline for Evaluating Vision Encoders in Multimodal Large Language Models}

\author{
  \!Yilin Yang$^{*\,1}$,
  Jun-Tao Tang$^{*\,2}$,
  Kengyi Wang$^{3}$,
  Siyuan Su$^{3}$,
  Gaoyong Luo$^{4}$,
  Mingda Chen$^{\dagger\,1}$\\
  {$^1$}School of Artificial Intelligence, Shanghai Jiao Tong University\\
  {$^2$}Nanjing University;
  {$^3$}Fudan University;
  {$^4$}Independent Researcher\\
  \footnotesize{\texttt{
  \{yilin.yang,mingdachen\}@sjtu.edu.cn;
  juntao.tang@smail.nju.edu.cn;}}\\
  \footnotesize{\texttt{
  \{24300980033,24300810015\}@m.fudan.edu.cn;
  210995214@mail.dhu.edu.cn}}
}

\newcommand{\momo}{RAVEL\xspace}
\newcommand{\ravel}{RAVEL\xspace}

\iclrfinalcopy 
\begin{document}

\maketitle
\begingroup
\renewcommand{\thefootnote}{\fnsymbol{footnote}}
\footnotetext[1]{Equal contribution.\qquad
\textsuperscript{\(\dagger\)} Corresponding author.}
\endgroup
\begin{abstract}

Evaluating vision encoders requires metrics that reliably predict their downstream performance in multimodal large language models (MLLMs). Although recent studies have shown that cross-modal metrics can better capture such performance, unimodal metrics remain the dominant choice in practice. In this work, we revisit cross-modal evaluation of vision encoders through large-scale experiments. We identify important limitations in both the experimental design and methodological formulation of prior approaches. After addressing these limitations and introducing simple improvements, we propose \ravel,\footnote{\textbf{R}etrieval-based
\textbf{A}ssessment of \textbf{V}ision \textbf{E}ncoders for \textbf{L}anguage Models} a training-free method based on cross-modal nearest-neighbor retrieval. Despite its simplicity, \ravel achieves state-of-the-art performance across our experiments, outperforming prior methods by a substantial margin. Our results demonstrate that simple cross-modal metrics, when evaluated under a careful and comprehensive setup, can provide a strong basis for evaluating vision encoders for MLLMs.\footnote{The code and corresponding checkpoints are available at
\url{https://github.com/JuntaoTang/MLLM-VisionEncoder-Eval}.}

\end{abstract}

\input{sections/intro}

\input{sections/relatedwork}
\input{sections/method_v2}
\input{sections/experimentsetups}

\input{sections/experimentresults}

\input{sections/conclusion}

\newpage
\subsection*{AI use statement}
In this work, we use generative AI tools for polishing the language of this paper, including improving grammar, clarity, and readability. We have not used generative AI tools for generating research ideas, develop the methodology, conduct experiments, analyze or interpret results, or formulate scientific claims. We have reviewed all AI-assisted edits. We take responsibility for the final content of this work, including text, claims or artifacts produced with the aid of generative AI.

\subsection*{Ethics Statement}
This work does not involve human subjects or the collection of new personal
data. All experiments are conducted using publicly available datasets and
pretrained models. We do not foresee ethical concerns specific to the proposed
method beyond those generally associated with MLLMs. In particular, the
evaluated models and datasets may contain biases or inappropriate content, and
the reported results should not be interpreted as guarantees of fairness or
safety. We encourage appropriate evaluation before applying the selected
models in real-world or high-stakes settings.

\subsection*{Reproducibility statement}
We provide detailed descriptions of the method, experimental protocol, and evaluation procedure to support reproducibility. Section~\ref{sec:method} formally defines RAVEL, including PCA-conditioned representations, patch-set visual relations, and cross-modal neighborhood agreement. Appendix~\ref{app:momo_algorithm} provides complete pseudocode. Section~\ref{sec:exp_overview} specifies the 70 vision encoders, three language backbones, paired calibration data, feature-extraction layers, and default hyperparameters used in our experiments.

All 210 MLLMs are trained using the same two-stage LLaVA-style pipeline. Appendix~\ref{app:model} documents the training datasets, image preprocessing, optimization settings, batch sizes, learning rates, random seeds, decoding configuration, benchmark subsets, and scoring rules. Appendix~\ref{APP:overview} reports the complete per-benchmark results for all encoder–language-model combinations. Appendix~\ref{app:detailed_mllm_results} defines the correlation metrics and FLOP accounting procedure, while Appendices~\ref{app:training_evaluation_details} and~\ref{app:implementation_details} provide the diagnostic analyses and implementation details for the evaluated baseline. Code, model configurations, evaluation splits, and extracted reference scores will be released upon publication.

\bibliography{iclr2027_conference}
\bibliographystyle{iclr2027_conference}

\newpage
\appendix
\input{sections/appendix}
\end{document}

%% file: math_commands.tex
\usepackage{amsmath,amsfonts,bm}

\def\eqref#1{equation~\ref{#1}}

\def\1{\bm{1}}

\DeclareMathAlphabet{\mathsfit}{\encodingdefault}{\sfdefault}{m}{sl}
\SetMathAlphabet{\mathsfit}{bold}{\encodingdefault}{\sfdefault}{bx}{n}



%% file: sections/intro.tex
\section{Introduction}

Recent multimodal large language models (MLLMs) have achieved strong performance in visual understanding and reasoning by using vision encoders to transform visual inputs into representations that can be processed by large language models \citep[\emph{inter alia}]{alayrac2022flamingo,llava}. The vision encoder\footnote{We use vision encoder as a general term for both continuous encoders and discrete visual tokenizers.} plays a critical role in this process: it determines what visual information is preserved and made available to the language model, and therefore substantially affect downstream MLLM performance. Reliably evaluating vision encoders and predicting their downstream performance in MLLMs is thus an important problem.

Current studies commonly evaluate vision encoders by integrating each candidate into an MLLM, training it with a unified multimodal recipe, and comparing the resulting models on downstream benchmarks~\citep{tong2024cambrian,cocchi2025llavamore,li2025openvision}. While this directly measures downstream utility, it requires a separate training and evaluation cycle for every encoder-language-model pair, making it prohibitively expensive as the candidate pool grows. Alternative approaches that seek to reduce computational cost instead rely on metrics that measure the quality of vision encoders in isolation, such as parameter count, zero-shot accuracy, or intrinsic representation statistics~\citep{radford2021clip,li2024bigger,garrido2023rankme}. However, these metrics do not account for how well a vision encoder is aligned or compatible with a particular language model. This motivates cross-modal metrics that directly compare visual and language representations without requiring expensive multimodal training or finetuning.

\begin{figure}[t]
\centering\includegraphics[width=0.5\linewidth]{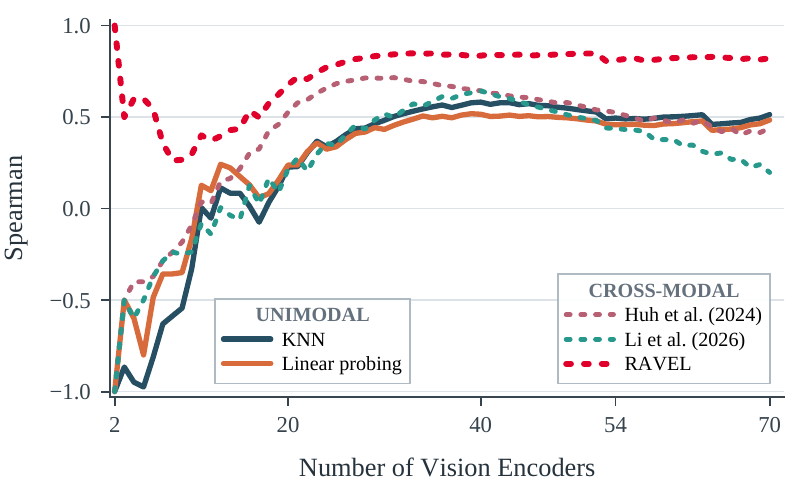}
  \caption{Effect of vision encoder candidate-pool size and heterogeneity on evaluation metrics. Cross-modal metrics outperform unimodal alternatives on moderately sized candidate pools, but their predictive performance drops substantially as more diverse vision encoders are added. In contrast, \ravel{}, with improved representation conditioning and finer-grained similarity scoring, consistently outperforms competing metrics across different vision encoder pools.}
  \label{fig:pool_sensitivity}
  \vspace{-5mm}
\end{figure}

Recent work approaches this problem by measuring similarities between visual and textual representation spaces. Notably, \citet{huh2024platonic} propose MutualNN, which retrieves the top-$k$ nearest neighbors independently in the visual and textual spaces for each paired sample and measures the overlap between their identities. Building on this idea, \citet{li2026rethinking} propose using Gromov--Wasserstein (GW) distance to capture more global structural similarity between the two representation distributions. While these cross-modal metrics are effective, our large-scale experiments reveal that their performance is highly sensitive to the composition of the candidate pool. Figure~\ref{fig:pool_sensitivity} reports the Spearman correlation between each metric's predictions and the ground-truth downstream performance of MLLMs as progressively more vision encoders are added to the candidate pool. Prior cross-modal metrics outperform unimodal alternatives on moderately sized pools, but their correlations drop substantially as the pool becomes larger and more diverse. In these more challenging settings, unimodal metrics again become stronger predictors of downstream MLLM performance. This sensitivity to the candidate pool may help explain why unimodal metrics remain widely used despite the promise of cross-modal evaluation.

To address this challenge, we revisit two fundamental aspects of representational similarity measurement. First, vector similarities can be strongly influenced by representation geometry. Features may be dominated by a few high-variance directions or exhibit strong correlations across dimensions, which can distort similarity measurements without reflecting downstream compatibility with the language model~\citep{garrido2023rankme,sasaki2023whitening}. We therefore build on PCA whitening, but modify the standard formulation with a stabilization term that prevents low-variance directions from being excessively amplified. Second, the granularity at which similarity is measured matters. Globally pooling visual features can discard fine-grained information contained in patch representations. We therefore introduce fine-grained patch-level similarity scoring into vision encoder evaluation, which, to our knowledge, has not been explored in prior work on this problem.

Motivated by these observations, we build upon MutualNN because it provides a simple and effective framework for measuring cross-modal similarity. We propose \ravel{}, a training-free cross-modal metric that retains its retrieval-based formulation while incorporating stabilized PCA whitening and fine-grained patch-level scoring.

We conduct a large-scale evaluation of nine strong metrics, covering both unimodal and cross-modal approaches, on 210 MLLMs comprising 70 vision encoders and three language models. This evaluation is substantially larger and more diverse than those in prior studies, allowing us to assess metric robustness across encoder architectures, pretraining objectives, representation types, and computational scales. Under FLOPs-controlled comparisons, \ravel{} improves over MutualNN by more than 42.8\% and outperforms the strongest prior metric by more than 21.1\% in correlation with ground-truth downstream MLLM performance. Further analysis shows that stabilized PCA whitening is critical to this improvement, while \ravel{} remains robust across a wide range of hyperparameter settings and across different vision encoder families. \ravel{} also outperforms TokBench \citep{wu2025tokbench}, a reconstruction-based evaluation benchmark specifically designed for reconstruction-trained vision encoders, demonstrating that its effectiveness extends beyond any single encoder family or training paradigm.

Our contributions can be summarized as follows.
\begin{itemize}
    \item We benchmark nine evaluation metrics on 210 MLLMs built from 70 vision encoders and three language models, revealing that existing cross-modal metrics degrade as encoder diversity increases.
    We will open-source our evaluation pipeline and model checkpoints to support future research.
\item We introduce \ravel{}, a simple training-free metric that combines stabilized PCA whitening with fine-grained patch-level similarity scoring.
\item Despite its simplicity, \ravel{} consistently outperforms prior methods and provides a strong baseline for future work on vision encoder evaluation.
\end{itemize}

%% file: sections/relatedwork.tex
\section{Related Work}

\noindent\textbf{Vision Encoders.}
Vision encoders can be broadly divided into continuous and discrete encoders. Continuous encoders produce real-valued global or patch-level representations and are commonly pretrained through vision--language learning~\citep{radford2021clip}, image-only self-supervision~\citep{oquab2023dinov2}, reconstruction~\citep{he2022mae}, or knowledge distillation~\citep{zhu2026eupe}. Discrete encoders quantize latent features into discrete codes using
explicit codebooks~\citep{oord2017vqvae}, lookup-free quantization, or binary quantization, yielding sequences of visual tokens. While often trained primarily for reconstruction, recent methods~\citep{ma2025unitok,lin2025toklip} also incorporate image--text contrastive objectives to preserve high-level semantics. Since both types determine what visual information is available to the language model, we evaluate them within a unified framework.

\noindent\textbf{Vision Encoder Evaluation and Selection.}
Prior work has explored cheaper alternatives to fully training an MLLM for every candidate encoder. Reconstruction-based benchmarks such as TokBench~\citep{wu2025tokbench} measure how well reconstructed images preserve fine-grained content. However, such methods require a reconstruction pathway and are therefore limited to encoders trained with reconstruction objectives.

Representation-based approaches apply more broadly. \citet{yang2025law} predict downstream MLLM performance from cross-modal alignment and visual correspondence, but require projector training, keypoint annotations, and downstream scores from a subset of fully trained MLLMs. \citet{zhang2025alignment} instead train a lightweight linear alignment layer between frozen vision and language backbones and evaluate it through zero-shot image--text retrieval.

Training-free methods directly compare visual and textual representation spaces. MutualNN~\citep{huh2024platonic} constructs $k$-nearest-neighbor sets independently in the two spaces and measures their overlap over paired samples. More recently, \citet{li2026rethinking} use Gromov--Wasserstein (GW) distance to compare their global geometry. MutualNN therefore emphasizes correspondence-aware local structure, whereas GW captures global structural similarity without directly using the known image--text pairings.

%% file: sections/method_v2.tex
\section{\ravel}
\label{sec:method}

\ravel{} builds on MutualNN. Given a paired image--text dataset, MutualNN first retrieves the top-$k$ nearest neighbors of each sample independently in the visual and textual representation spaces. It then uses the known image--text correspondences to measure the overlap between the two sets of retrieved samples. Intuitively, if a vision encoder is compatible with a target language model, images that are close in the visual representation space should have corresponding texts that are also close in the language model's representation space.

\ravel{} retains this retrieval-and-overlap pipeline, but revisits how similarities are measured within the two representation spaces. In particular, we introduce two modifications: stabilized PCA whitening to reduce representation-specific geometric bias, and patch-level scoring functions to preserve fine-grained visual similarities.

\subsection{Preliminary: MutualNN}
\label{sec:relation}

Let $\mathcal{D}=\{(x_i,t_i)\}_{i=1}^{n}$ denote a dataset of paired images and texts. Given a candidate vision encoder $E$ and a target language model $L$, MutualNN represents each image and text with a single vector:
\begin{equation}
\mathbf{z}_i^{v}=E(x_i),
\qquad
\mathbf{z}_i^{t}=L(t_i).
\end{equation}
Visual and textual representations may have different dimensions and reside in different feature spaces, so MutualNN does not compare them directly. Instead, it independently computes pairwise similarities within the two representation spaces:
\begin{equation}
s_{ij}^{v}
=
\operatorname{sim}
\left(
\mathbf{z}_i^{v},
\mathbf{z}_j^{v}
\right),
\qquad
s_{ij}^{t}
=
\operatorname{sim}
\left(
\mathbf{z}_i^{t},
\mathbf{z}_j^{t}
\right).
\label{eq:within-view-relations}
\end{equation}
For each sample $i$, the top-$k$ nearest neighbors are then retrieved independently in the visual and textual spaces:
\begin{equation}
\mathcal{N}^{v}(i)
=
\operatorname*{arg\,topk}_{j\neq i}
s_{ij}^{v},
\qquad
\mathcal{N}^{t}(i)
=
\operatorname*{arg\,topk}_{j\neq i}
s_{ij}^{t}.
\label{eq:retrieval-neighborhoods}
\end{equation}
Because $x_j$ and $t_j$ correspond to the same underlying sample, the retrieved identities can be directly compared across modalities. MutualNN measures their average overlap as
\begin{equation}
S
=
\frac{1}{n}
\sum_{i=1}^{n}
\frac{
\left|
\mathcal{N}^{v}(i)
\cap
\mathcal{N}^{t}(i)
\right|
}{k}.
\label{eq:retrieval-agreement}
\end{equation}
A higher score indicates that the vision encoder and language model induce more similar neighborhood structures over the paired dataset. \ravel{} retains this scoring procedure, but modifies how the similarities used to construct these neighborhoods are computed.

\subsection{Stabilized PCA Whitening}
\label{sec:pca-relations}

The first modification concerns the geometry of the representation spaces. Vector similarities can be strongly affected by the variance and correlation structure of the underlying features. PCA whitening has long been used to mitigate this problem in retrieval. In image retrieval, \citet{jegou2012negative} show that whitening can reduce the influence of common, correlated co-occurrence directions, and related work applies PCA whitening to deep image representations for retrieval~\citep{babenko2015aggregating}. For text, whitening has likewise been used to improve sentence representations and semantic retrieval~\citep{su2021whitening,huang-etal-2021-whiteningbert-easy}.

We therefore apply PCA whitening before measuring representation similarity.

While whitening suppresses dominant high-variance directions, we find that directly scaling each principal direction by the inverse square root of its variance can excessively amplify low-variance directions, which may primarily capture noise. We therefore introduce a small stabilization term $\epsilon=10^{-4}$ to the PCA eigenvalues and apply the following transformation to a particular encoded text or image vector, $\mathbf{z}_i$:
\begin{equation}
\left(
\boldsymbol{\Lambda}
+
\epsilon\mathbf{I}
\right)^{-1/2}
\mathbf{U}^{\top}
\left(
\mathbf{z}_i-\boldsymbol{\mu}
\right)
\label{eq:pca-whitening}
\end{equation}
Where $\mathbf U$ contains the eigenvectors associated with the positive eigenvalues of the covariance matrix, and $\boldsymbol{\Lambda}$ is the diagonal matrix of those eigenvalues. $\mathbf{\mu}$ is the mean of the given dataset. The stabilization term limits the amplification of low-variance directions while retaining the ability of PCA whitening to suppress dominant directions. We fit the transformation independently for each candidate vision encoder and for the target language model using only representations encoded from $\mathcal{D}$. For the visual space, PCA is estimated over individual patch representations. For the textual space, PCA is estimated over mean-pooled textual representations. No downstream labels or MLLM training results are used.

\subsection{Fine-Grained Similarity Scoring}
\label{sec:patch-relations}

The second modification concerns the granularity at which similarity is measured. Both visual and textual representations can be viewed as sets of local features, such as image patches or text tokens, and global pooling may discard useful fine-grained correspondences. However, the appropriate granularity differs across modalities in our setting. Vision encoders typically produce hundreds of patch tokens per image, whereas the paired textual descriptions are much shorter. Unlike text retrieval methods such as ColBERT \citep{khattab2020colbert}, which operate on relatively long passages and documents, our textual sequences are short enough that global representations are sufficient, while fine-grained matching is more important for vision.

For visual representations, we seek a similarity function that directly compares two sets of patch features. Chamfer-style similarity is well suited to this setting because it is permutation-invariant, naturally handles sets with different numbers of elements, and captures local correspondences without requiring a predefined one-to-one alignment. By matching each feature to its most similar counterpart, it can preserve strong local agreements that may be diluted by global pooling. Chamfer-style matching has been used in set-based problems such as point-cloud matching and reconstruction~\citep{Fan_2017_CVPR} and fine-grained video similarity~\citep{Kordopatis-Zilos_2019_ICCV}. Following the same set-matching principle, we apply it directly to patch representations.

We define similarity between two images using a symmetric Chamfer-style scoring function:
\begin{equation}
\begin{aligned}
s_{ij}^{v}
=
\frac{1}{2}
\Bigg[
&
\frac{1}{M_i}
\sum_{a=1}^{M_i}
\max_{1\leq b\leq M_j}
\operatorname{sim}
\left(
\widetilde{\mathbf{p}}_{ia},
\widetilde{\mathbf{p}}_{jb}
\right)
+
\frac{1}{M_j}
\sum_{b=1}^{M_j}
\max_{1\leq a\leq M_i}
\operatorname{sim}
\left(
\widetilde{\mathbf{p}}_{jb},
\widetilde{\mathbf{p}}_{ia}
\right)
\Bigg],
\end{aligned}
\label{eq:patch-relation}
\end{equation}
where $M_i$ and $M_j$ denote the numbers of patch tokens in images $i$ and $j$, and $\widetilde{\mathbf{p}}_{ia}$ is the $a$-th patch representation of image $i$. The two terms perform matching in opposite directions, producing a symmetric image-level similarity. Special tokens, e.g., CLS tokens, are excluded.

For text, let $\widetilde{\mathbf{z}}_i^{t}$ denote the mean-pooled representation of caption $t_i$, and compute textual similarity as $s_{ij}^{t}=\operatorname{sim}(\widetilde{\mathbf{z}}_i^{t},\widetilde{\mathbf{z}}_j^{t})$. The visual and textual similarities are used to construct the nearest-neighbor sets in Equation~\ref{eq:retrieval-neighborhoods}, and the final \ravel{} score is computed using Equation~\ref{eq:retrieval-agreement}.

%% file: sections/experimentsetups.tex
\section{Experiments}
\subsection{Experimental Setups}
\noindent\textbf{Overall Pipeline.}
\label{sec:exp_overview}
Our experiments comprise two stages. First, we establish the ground-truth downstream performance for each vision encoder--language-model pair $(E,L)$. We train and evaluate MLLMs on a collection of downstream benchmarks. Let
\begin{equation}
\mathbf{P}(E,L)
=
\left[
P_1(E,L),\ldots,P_B(E,L)
\right]
\end{equation}
denote its performance across $B$ benchmarks. We use the average benchmark performance
\begin{equation}
P(E,L)
=
\frac{1}{B}
\sum_{b=1}^{B}
P_b(E,L)
\end{equation}
as the ground-truth performance of the pair. Repeating this procedure across different vision encoders and language models gives a collection of tuples $(E,L,P)$.

Second, for each evaluation metric, we compute a score $S(E,L)$ that predicts the compatibility between vision encoder $E$ and language model $L$ with minimal training of the corresponding MLLM. We assess each metric by measuring Pearson and Spearman correlations between its predicted scores $S(E,L)$ and the ground-truth downstream performance $P(E,L)$ across vision encoder--language-model pairs.
More details on the evaluation metrics are in Appendix~\ref{app:detailed_metric}.

When training MLLMs, we continue training from existing pretrained language model checkpoints and perform mid-training on LLaVA-LCS-558K~\citep{liu2024llava15}, optimizing only the projector while keeping both the vision encoder and language model backbone frozen. We then jointly optimize the projector and language model backbone on LLaVA-665K~\citep{liu2024llava15}.
The resulting MLLMs are evaluated on 11 benchmarks spanning visual question
answering, multimodal reasoning, OCR, captioning, and hallucination assessment.
The unweighted average across the benchmarks defines the ground-truth performance. Details are provided in Appendix~\ref{app:training_evaluation_details}. 

\noindent\textbf{Evaluated Models.}
We evaluate a pool of 70 vision encoders spanning continuous encoders and discrete visual tokenizers. Each vision encoder is evaluated with Qwen3-1.7B~\citep{yang2025qwen3}, Qwen2.5-1.5B-Instruct~\citep{yang2024qwen25}, and SmolLM2-1.7B-Instruct~\citep{allal2025smollm2}. Further details on the candidate vision encoders are provided in
Appendix~\ref{app:model}.

\noindent\textbf{Baselines.}
We compare \ravel{} against vision-only metrics that ignore the target language model, namely linear probing~\citep{caron2021dino} and $k$-nearest-neighbor classification (KNN; \citealp{wu2018unsupervised}), and against training-free cross-modal metrics: RSA~\citep{kriegeskorte2008rsa}, CCA~\citep{hotelling1936cca}, GW~\citep{li2026rethinking}, and MutualNN~\citep{huh2024platonic}. As stronger but costlier reference points, we also include methods that require additional training or fitting: alignment probing~\citep{zhang2025alignment}, AC Policy~\citep{yang2025law}, and mid-training loss, i.e., the final projector loss of FLOP-matched first-stage mid-training. Details are in Appendix~\ref{app:baseline}.

%% file: sections/experimentresults.tex
\begin{table*}[t]
\centering
\small
\setlength{\extrarowheight}{1pt}
\setlength{\tabcolsep}{4pt}
\renewcommand{\arraystretch}{1.0}
\caption{
FLOPs-controlled comparison of different evaluation metrics.
All methods use matched computational budgets except AC Policy$^\dagger$, whose computational cost cannot be reduced to the matched FLOP budget due to its method design.
Best and second-best results are shown in {bold} and underlined,
respectively.
}
\label{tab:main_results}
{
\begin{tabularx}{\textwidth}{
    >{\raggedright\arraybackslash}p{0.4\textwidth}|
    >{\centering\arraybackslash}X
    >{\centering\arraybackslash}X|
    >{\centering\arraybackslash}X
    >{\centering\arraybackslash}X|
    >{\centering\arraybackslash}X
    >{\centering\arraybackslash}X
}
\hline
\rowcolor{gray!20}
&
\multicolumn{2}{c|}{\textbf{Qwen3-1.7B}} &
\multicolumn{2}{c|}{\textbf{Qwen2.5-1.5B}} &
\multicolumn{2}{c}{\textbf{SmolLM2-1.7B}} \\
\cline{2-7}

\rowcolor{gray!20}
\textbf{Method} &
$\rho\uparrow$ & $r\uparrow$ &
$\rho\uparrow$ & $r\uparrow$ &
$\rho\uparrow$ & $r\uparrow$ \\
\hline

\rowcolor{gray!10}
\multicolumn{7}{l}{\textit{Vision-only metrics}} \\

Linear probing~\citep{caron2021dino}
& 0.484 & 0.490
& 0.568 & 0.510
& 0.593 & 0.597 \\

KNN~\citep{wu2018unsupervised}
& 0.551 & 0.513 
& 0.534 & 0.598 
& 0.603 & 0.567\\

\hline
\rowcolor{gray!10}
\multicolumn{7}{l}{\textit{Cross-modal metrics requiring additional model training}} \\

Mid-Training Loss
& 0.342 & 0.349
& 0.434 & 0.403
& 0.272 & 0.300 \\

AC Policy$^\dagger$~\citep{yang2025law}
& 0.506 & 0.445
& 0.573 & 0.391
& 0.632 & 0.603 \\

Alignment probing~\citep{zhang2025alignment}
& \underline{0.569} & \underline{0.618}
& \underline{0.634} & \underline{0.640}
& \underline{0.670} & \underline{0.711} \\

\hline
\rowcolor{gray!10}
\multicolumn{7}{l}{\textit{Cross-modal metrics without requiring additional model training}} \\

RSA~\citep{kriegeskorte2008rsa}
& 0.125 & 0.206
& 0.015 & 0.068
& -0.070 & 0.060 \\

CCA~\citep{hotelling1936cca}
& 0.163 & 0.251
& 0.275 & 0.338
& 0.213 & 0.341 \\

MutualNN~\citep{huh2024platonic}
& 0.439 & 0.543
& 0.468 & 0.570
& 0.361 & 0.541 \\

GW~\citep{li2026rethinking}
& 0.226 & 0.236
& 0.515 & 0.312
& 0.577 & 0.472 \\

\rowcolor{ScoreMint!60}
{\momo{} (Ours)}
& \textbf{0.820} & \textbf{0.816}
& \textbf{0.897} & \textbf{0.896}
& \textbf{0.833} & \textbf{0.845} \\
\hline
\end{tabularx}
}
\vspace{-2mm}
\end{table*}

\subsection{Main Results}
\label{sec:main_results}
Table~\ref{tab:main_results} compares \momo{} with representative vision encoder evaluation methods. \momo{} achieves the highest correlations across all three language backbones, with Spearman correlations of \(0.820\), \(0.897\), and \(0.833\) and Pearson correlations of \(0.816\), \(0.896\), and \(0.845\) for Qwen3-1.7B, Qwen2.5-1.5B, and SmolLM2-1.7B, respectively. Compared with the strongest baseline, alignment probing, \momo{} improves Spearman correlation by \(0.251\), \(0.263\), and \(0.163\), and Pearson correlation by \(0.198\), \(0.256\), and \(0.134\). These consistent gains show that \momo{} more accurately predicts both the relative ranking of candidate encoders and how their downstream MLLM performance varies. 

\begin{table*}[t]
\centering

\begin{minipage}[t]{0.39\textwidth}
\vspace{0pt}
\centering

{\fontsize{8}{9.5}\selectfont
\renewcommand{\arraystretch}{1.10}
\setlength{\tabcolsep}{3pt}
\captionsetup{skip=4pt}

\begin{minipage}[t][6\baselineskip][t]{\linewidth}
    \captionof{table}{Comparing the effect of patch-level scoring functions and PCA whitening. The results are averaged over the three language models.}
    \label{ablation}
\end{minipage}

\begin{tabular}{@{}lcc@{}}
    \toprule
    \multirow{2}{*}{%
        \makebox[0.64\linewidth][l]{\textbf{Variant}}%
    }
        & \multicolumn{2}{c}{\textbf{Metrics}} \\
    \cmidrule(lr){2-3}
        & $\rho\uparrow$ & $r\uparrow$ \\
    \midrule
    MutualNN
        & 0.194 & 0.240 \\
    $+$ PCA Whitening
        & 0.782 & 0.798 \\
    $+$ Patch-level w/o PCA
        & 0.548 & 0.510 \\
    $+$ Patch-level w/ PCA
        & \textbf{0.850} & \textbf{0.852} \\
    \bottomrule
\end{tabular}
}

\end{minipage}%
\hspace{0.015\textwidth}%
\begin{minipage}[t]{0.58\textwidth}
\vspace{0pt}
\centering

{\fontsize{8}{9.5}\selectfont
\renewcommand{\arraystretch}{1.10}
\setlength{\tabcolsep}{2pt}
\captionsetup{skip=4pt}

\begin{minipage}[t][6\baselineskip][t]{\linewidth}
    \captionof{table}{Correlations of TokBench metrics and RAVEL with downstream MLLM performance using discrete vision encoders.}
    \label{tab:tokbench}
\end{minipage}

\begin{tabular}{@{}lcccccc@{}}
    \toprule
    \multirow{2}{*}{\textbf{Metric}}
        & \multicolumn{2}{c}{\textbf{Qwen3}}
        & \multicolumn{2}{c}{\textbf{Qwen2.5}}
        & \multicolumn{2}{c}{\textbf{SmolLM2}} \\
    \cmidrule(lr){2-3}
    \cmidrule(lr){4-5}
    \cmidrule(lr){6-7}
        & $\rho\uparrow$ & $r\uparrow$
        & $\rho\uparrow$ & $r\uparrow$
        & $\rho\uparrow$ & $r\uparrow$ \\
    \midrule
    T-ACC
        & $+0.359$ & $+0.364$
        & $-0.308$ & $+0.088$
        & $-0.975$ & $-0.650$ \\
    T-NED
        & $+0.359$ & $+0.255$
        & $-0.308$ & $-0.016$
        & $-0.975$ & $-0.679$ \\
    F-Sim
        & $+0.359$ & $-0.223$
        & $-0.308$ & $-0.433$
        & $-0.975$ & $-0.700$ \\
    RAVEL
        & $\mathbf{+0.600}$ & $\mathbf{+0.743}$
        & $\mathbf{+0.600}$ & $\mathbf{+0.598}$
        & $\mathbf{+0.400}$ & $\mathbf{+0.524}$ \\
    \bottomrule
\end{tabular}
}

\end{minipage}

\vspace{-3mm}
\end{table*}

\begin{figure*}[t]
    \centering
    \begin{subfigure}[t]{0.32\textwidth}
        \centering
        \includegraphics[width=\linewidth]{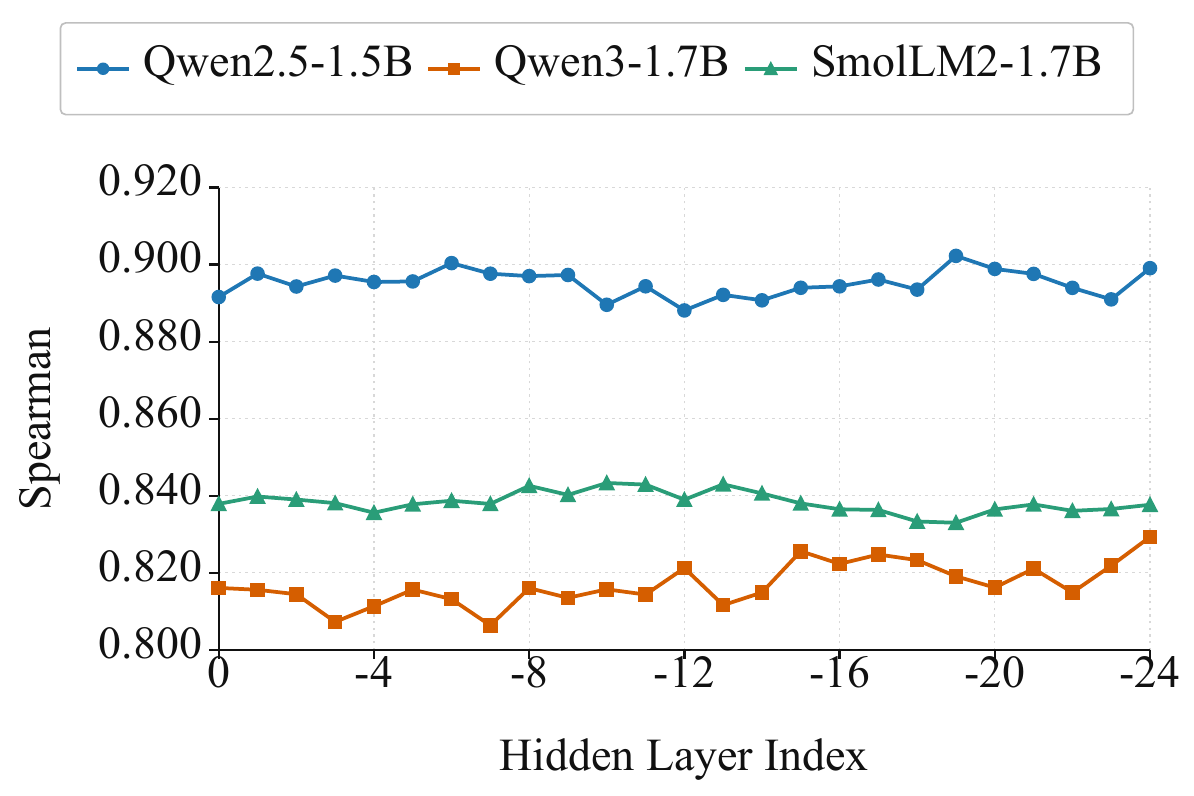}
        \caption{}
        \label{fig:layers}
    \end{subfigure}\hfill
    \begin{subfigure}[t]{0.32\textwidth}
        \centering
        \includegraphics[width=\linewidth]{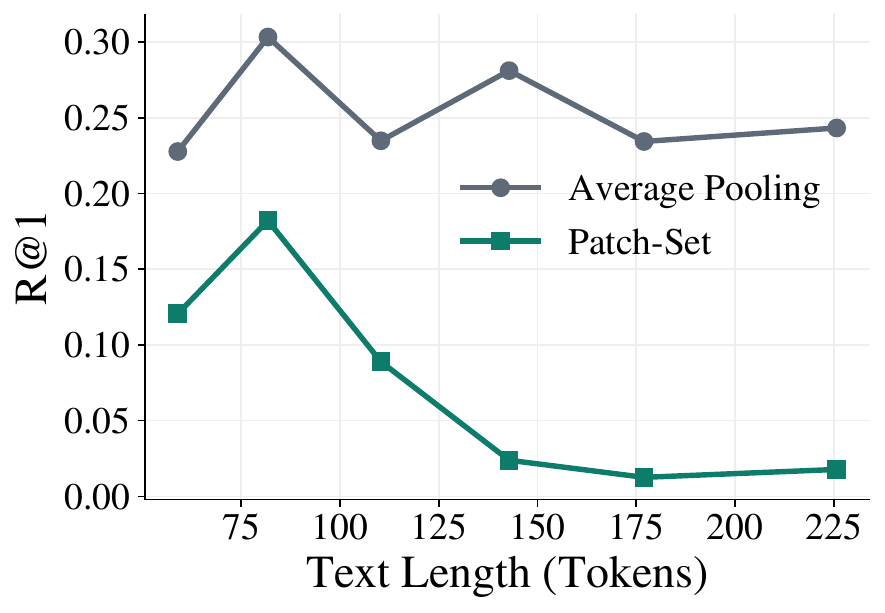}
        \caption{}
        \label{fig:multipatch}
    \end{subfigure}\hfill
    \begin{subfigure}[t]{0.32\textwidth}
        \centering
        \includegraphics[width=\linewidth]{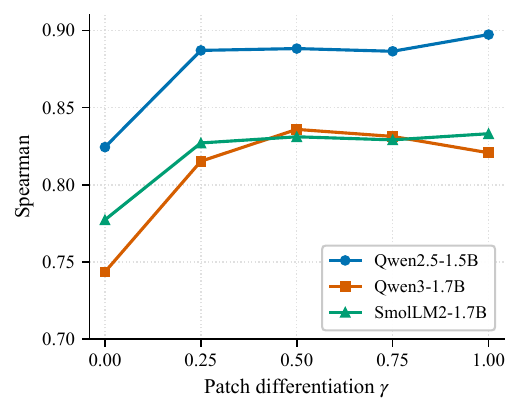}
        \caption{}
        \label{fig:patch-differentiation}
    \end{subfigure}
    \caption{{Feature representation and patch-set matching in \momo{}.}
    (a) Effect of the text feature extraction layer on ranking correlation. (b) Patch-set matching does not benefit long captions. (c) Patch differentiation under controlled feature intervention. }
    \label{fig:pca-patch-mechanisms}
\end{figure*}

\begin{figure*}[t]
    \centering
    \captionsetup[subfigure]{skip=2pt}

    \begin{subfigure}[t]{0.32\textwidth}
        \vspace{0pt}
        \centering
        \includegraphics[height=0.58\linewidth]{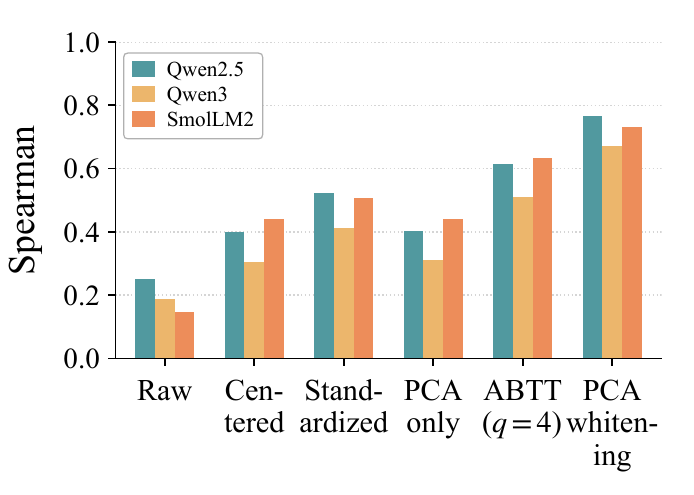}
        \caption{Post-processing methods}
        \label{fig:post_processing_methods}
    \end{subfigure}\hfill
    \begin{subfigure}[t]{0.32\textwidth}
        \vspace{0pt}
        \centering
        \includegraphics[height=0.58\linewidth]{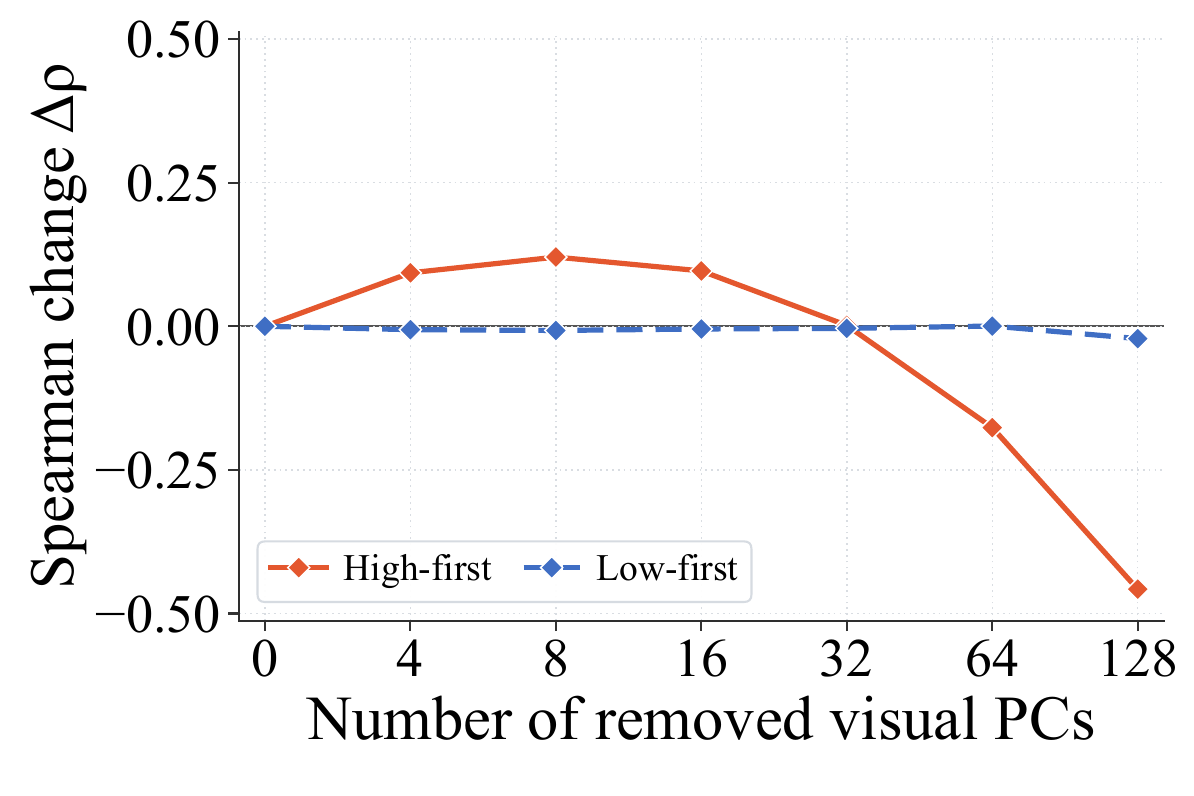}
        \caption{Sensitivity to visual PC removal}
        \label{fig:pca_sensitivity}
    \end{subfigure}\hfill
    \begin{subfigure}[t]{0.32\textwidth}
        \vspace{0pt}
        \centering
        \includegraphics[height=0.58\linewidth]{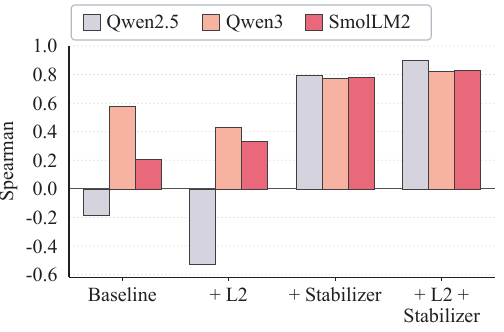}
        \caption{Ablation of PCA whitening}
        \label{fig:pca_ablation}
    \end{subfigure}

    \caption{{Representation post-processing choices.}
    (a) Five post-processing methods and the unprocessed baseline.
    (b) Spearman change after removing high- or low-variance visual PCs.
    (c) Ablation of L2 normalization and spectral stabilization under conventional PCA whitening.}
    \label{fig:pca-postprocessing-abtt}
    \vspace{-5mm}
\end{figure*}

\subsection{Empirical Analysis of \momo{}}

\subsubsection{Effect of Patch-Level Scoring Function and PCA Whitening}
\label{sec:abalation}
Table~\ref{ablation} reports ablation results averaged over the three language backbones. Starting from cross-modal nearest-neighbor retrieval ($k = 100$) with globally pooled visual representations, stabilized PCA whitening improves Spearman's $\rho$ from $0.194$ to $0.782$ and Pearson's $r$ from $0.240$ to $0.798$. Replacing global visual similarity with patch-level Chamfer-style scoring further increases the correlations to $0.850$ and $0.852$, respectively, demonstrating the complementary benefits of geometric conditioning and fine-grained visual matching.

\subsubsection{Analysis of Text Representation}
\noindent\textbf{Textual Embedding Layer Selection.}
Unlike vision encoders, which are explicitly trained to produce visual representations, language models do not provide a uniquely defined embedding layer. Their hidden states may encode different levels of semantic information across layers, making the choice of text feature extraction layer an additional design decision. We therefore examine how this choice affects \ravel{}. As shown in Figure~\ref{fig:layers}, Spearman correlations vary only slightly across layers. Given this limited sensitivity, we follow the setting of \citealp{li2026rethinking} and use the second-to-last hidden layer throughout.

\noindent\textbf{Do We Need Patch-Level Scoring Functions for Textual Representations?}
To test whether fine-grained token matching also benefits textual similarity estimation, we represent each text as a set of content-token embeddings and compute similarity using ColBERT-style MaxSim~\citep{khattab2020colbert}, which averages the maximum token-wise similarity for each query token. On a same-image caption retrieval task constructed from CC3M, mean pooling consistently outperforms this token-level formulation, with the performance gap widening as caption length increases, as shown in Figure~\ref{fig:multipatch}. We therefore use mean-pooled token representations for textual similarity.

\subsubsection{Analysis of PCA Whitening}
\noindent\textbf{Why PCA Whitening?}
Figure~\ref{fig:post_processing_methods} shows that PCA whitening achieves the highest average rank correlation among the evaluated post-processing methods, with details of each method provided in the appendix~\ref{app:postprocessing-methods}. Following prior dimension-removal analyses~\citep{timkey2021allbark}, We further examine the effect of explicit component removal by varying the number of leading PCs removed by ABTT and trailing PCs removed before whitening. Figure~\ref{fig:pca_sensitivity} shows that ABTT benefits from removing a few leading PCs, but these gains diminish and eventually reverse as more components are removed. For PCA whitening, removing low-variance PCs provides no clear improvement over the no-removal setting within the tested range. We therefore omit this additional truncation step and retain all positive-eigenvalue directions.

\noindent\textbf{Stabilizing PCA Whitening.}
We further examine the implementation choices required for reliable whitening. In Figure~\ref{fig:pca_ablation}, the baseline denotes conventional PCA whitening. The results show that scale normalization alone is insufficient: the main improvement comes from stabilizing the small-eigenvalue directions, after which L2 normalization provides a further gain. We therefore use stabilized and normalized PCA whitening in \momo{}.

\noindent\textbf{Effects Across Methods.}
Table~\ref{tab:pca_effect} further shows that this post-processing benefits all evaluated methods, although the improvement is less pronounced for GW. Unlike the other methods, GW evaluates the complete distance geometry and applies median-ratio scaling to align cross-modal distance scales~\citep{li2026rethinking}. While this adjustment corrects differences in overall magnitude, it does not align the shapes of the distance distributions. These results suggest that PCA whitening provides broadly useful representation conditioning, with particularly strong benefits for methods based on local or correlation-based comparisons. 

\begin{figure*}[t]
\centering
\begin{subfigure}[t]{0.4475\textwidth}
  \centering
  \vspace{0pt}
  \includegraphics[width=\linewidth]{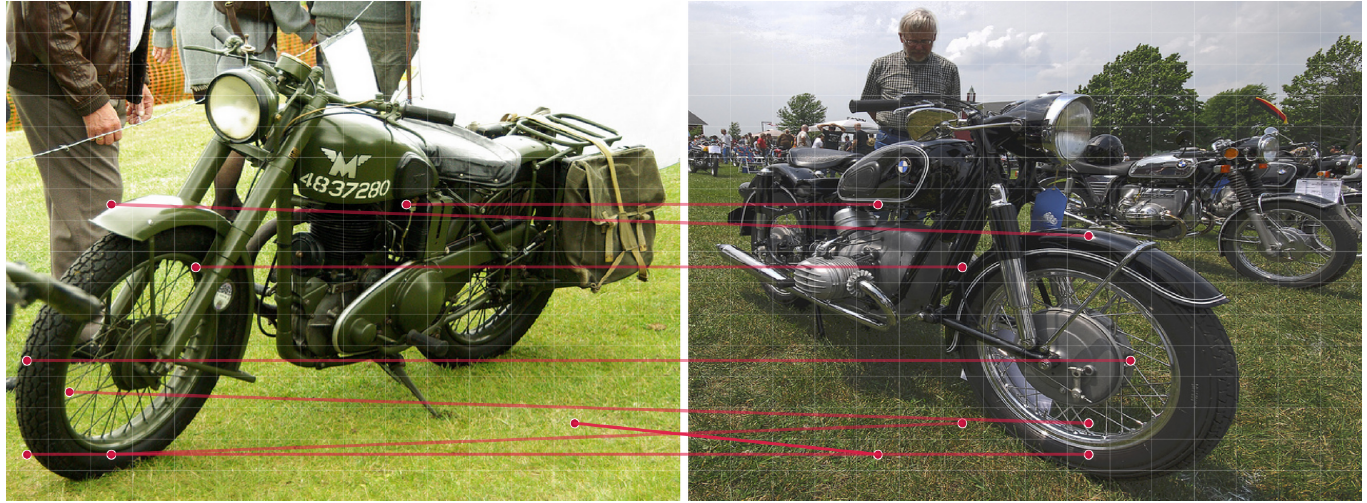}
  \label{fig:patch_correspondences}
  \caption{Patch correspondences.}
\end{subfigure}
\hfill
\begin{subfigure}[t]{0.5425\textwidth}
  \centering
  \vspace{0pt}
  \includegraphics[width=\linewidth]{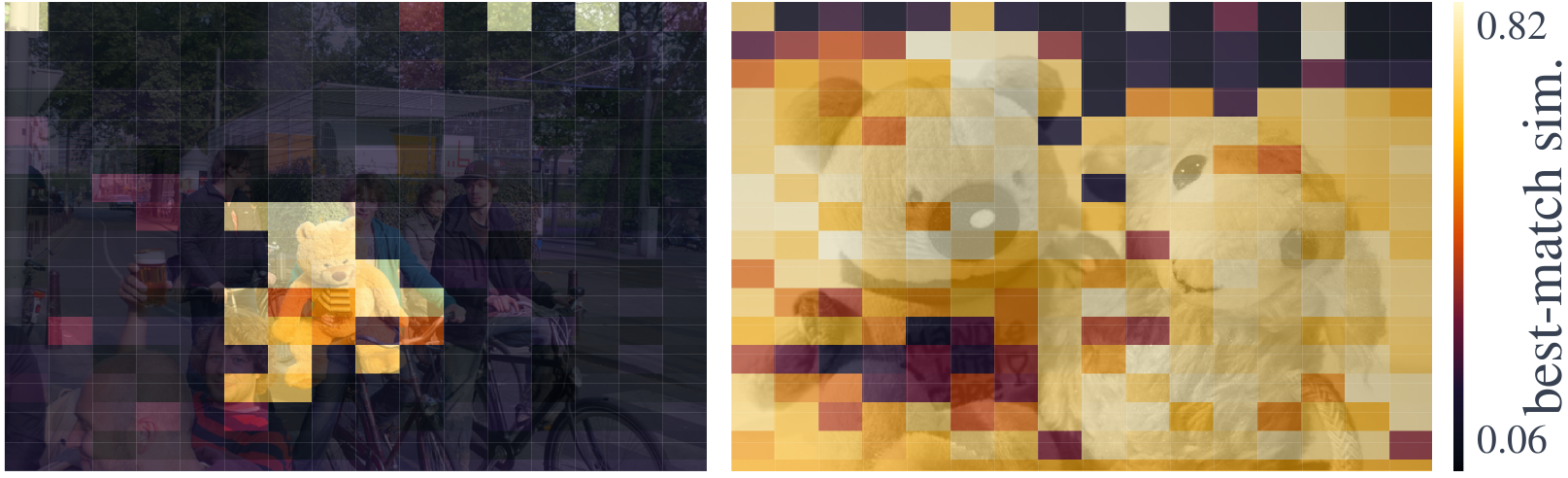}
  \label{fig:similarity_maps}
  \caption{Per-patch best-match similarity maps.}
\end{subfigure}
\caption{{Patch-level matching preserves local structure.}
(a)~Red lines link high-similarity patch pairs.
(b)~Color shows each patch's best-match similarity (brighter $=$ stronger).}
\label{fig:patch-mechanism}
\vspace{-3mm}
\end{figure*}

\subsubsection{Effects of Patch-Level Matching}
Global pooling compresses all visual tokens into a single vector, potentially mixing shared objects with unrelated context. Figure~\ref{fig:patch-mechanism} visualizes patch correspondences and per-patch best-match scores. The strongest matches concentrate on shared motorcycle regions despite changes in viewpoint and background, while unmatched regions receive lower scores, suggesting that patch matching can emphasize shared local semantics. To test whether this local information contributes to tokenizer ranking, we progressively shrink patch features toward their within-image mean while keeping the remaining pipeline fixed. Figure~\ref{fig:patch-differentiation} shows that the Spearman rank correlation across language backbones generally increases under both graph settings as $\gamma$ is increased from $0$ to $1$. This intervention provides independent evidence that preserving within-image patch differentiation improves tokenizer-ranking agreement in the tested settings.

\begin{table*}[t]
\centering
\begin{minipage}[t]{0.50\textwidth}
\vspace{0pt}
\centering
\captionsetup{justification=centering}
\captionof{table}[Hyperparameter sensitivity]{Hyperparameter sensitivity.}
\label{tab:hyperparameter_sensitivity}
\vspace{-2mm}

\begingroup
\small
\setlength{\tabcolsep}{3.2pt}
\renewcommand{\arraystretch}{1.21}
\resizebox{\linewidth}{!}{%
\begin{tabular}{@{}c|ccc|ccc@{}}
\toprule
{\normalsize\bfseries Param.}
& \multicolumn{3}{c|}{{\normalsize $k$}}
& \multicolumn{3}{c}{{\normalsize $\epsilon_w\;(\times10^{-4})$}} \\
\cmidrule(lr){2-4}
\cmidrule(lr){5-7}

{\normalsize\bfseries Value}
& 50 & \textbf{100} & 150
& 0.5 & \textbf{1} & 5 \\
\midrule

{\normalsize\boldmath$\rho$}
& 0.838 & \textbf{0.850} & 0.850
& 0.829 & \textbf{0.850} & 0.846 \\

{\normalsize\boldmath$r$}
& 0.847 & \textbf{0.852} & 0.844
& 0.829 & \textbf{0.852} & 0.848 \\
\bottomrule
\end{tabular}
}
\endgroup
\end{minipage}%
\hspace{0.024\textwidth}%
\begin{minipage}[t]{0.45\textwidth}
\vspace{0pt}
\centering
\captionsetup{justification=centering}
\captionof{table}[Effect of PCA whitening]{PCA-whitening effect.}
\label{tab:pca_effect}
\vspace{-2mm}

\begingroup
\small
\setlength{\tabcolsep}{3pt}
\renewcommand{\arraystretch}{0.9}
\resizebox{\linewidth}{!}{%
\begin{tabular}{@{}lccc@{}}
\toprule
\textbf{Method}
& \textbf{Native $\rho$}
& \textbf{PCA-W $\rho$}
& $\boldsymbol{\Delta\rho}$ \\
\midrule
MutualNN & 0.422 & 0.749 & +0.327 \\
GW       & 0.439 & 0.497 & +0.058 \\
RSA      & 0.024 & 0.708 & +0.684 \\
CCA      & 0.217 & 0.448 & +0.231 \\
\bottomrule
\end{tabular}%
}
\endgroup
\end{minipage}
\vspace{-3mm}
\end{table*}

\subsection{Robustness and Computational Efficiency}
\label{app:robustness_efficiency}

\begin{figure}[t]
  \centering

  \includegraphics[width=\linewidth]
  {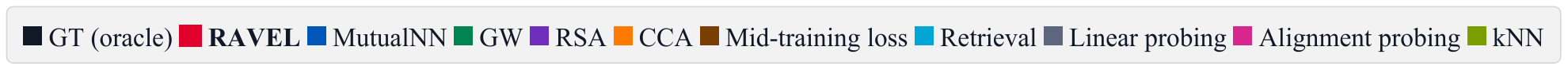}
  \par\vspace{0.35em}

  \begin{subfigure}[t]{0.33\linewidth}
    \centering
    \includegraphics[width=\linewidth]
    {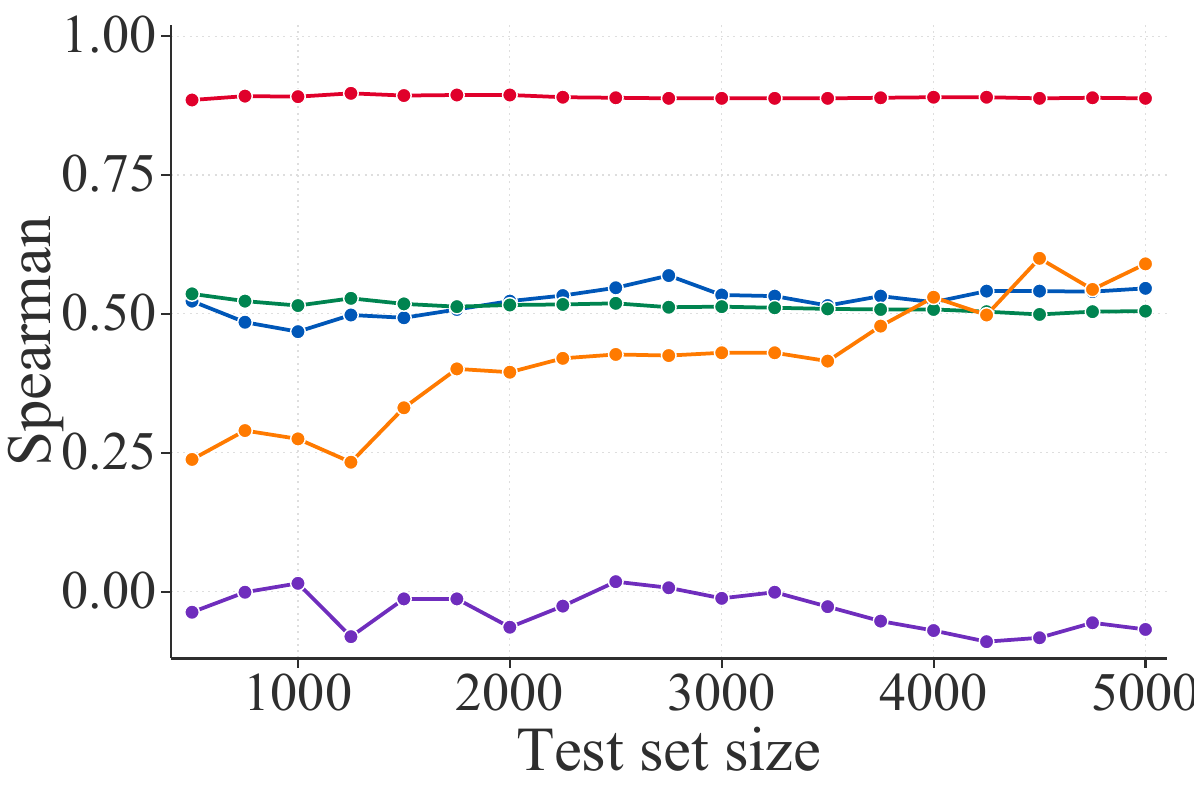}
    \caption{Test-set-size sensitivity.}
    \label{fig:robustness_test_size}
  \end{subfigure}%
  \hspace{0.025\linewidth}%
  \begin{subfigure}[t]{0.33\linewidth}
    \centering
    \includegraphics[width=\linewidth]
    {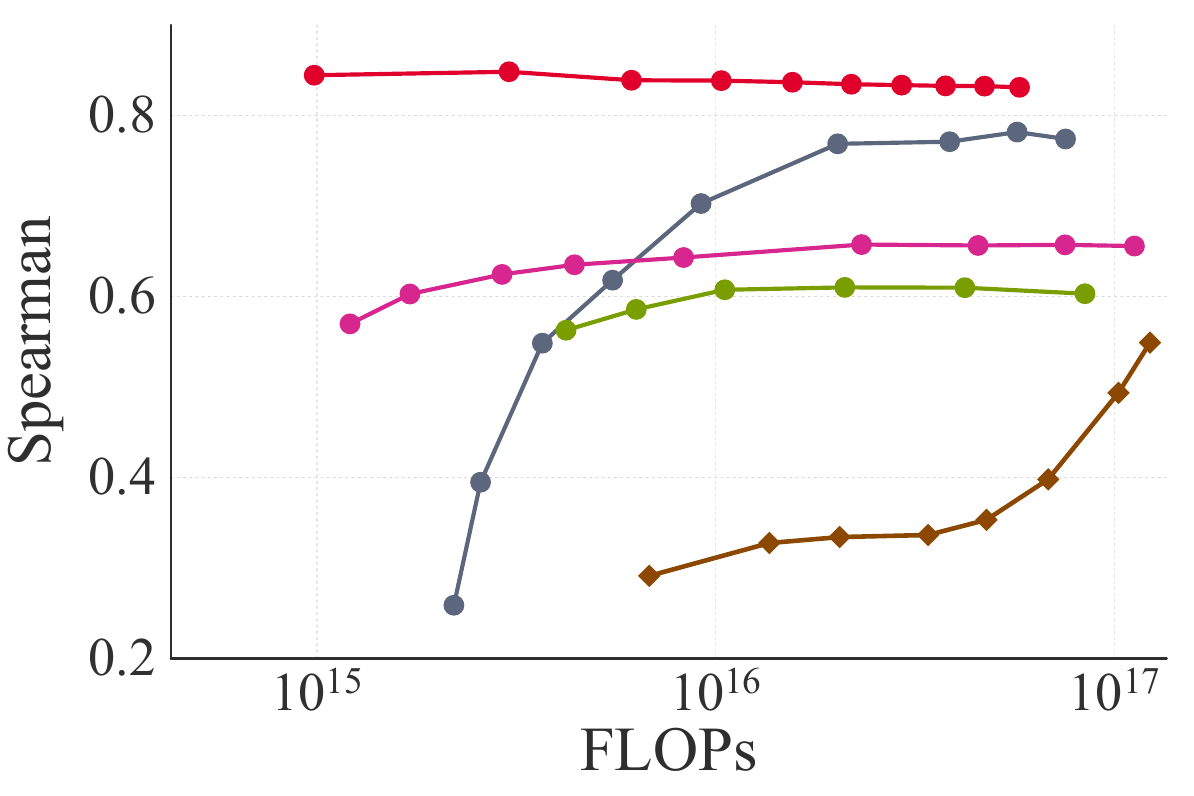}
    \caption{Accuracy--efficiency trade-off.}
    \label{fig:spearman_flops}
  \end{subfigure}

  \caption{Robustness and scaling behavior.}
  \label{fig:robustness_efficiency}
  \vspace{-1mm}
\end{figure}

\begin{figure}[t]
    \centering
    \includegraphics[
        width=\linewidth,
        height=0.20\textheight,
        keepaspectratio
    ]{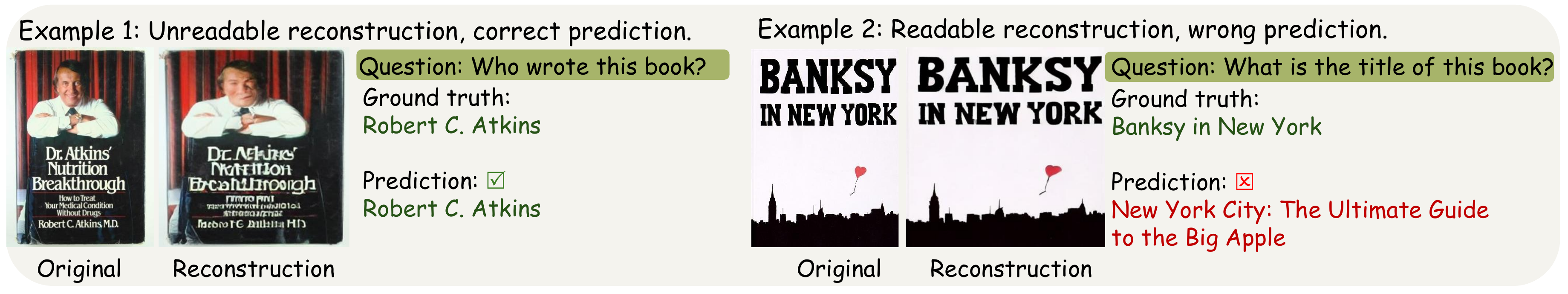}
    \caption{{Reconstruction quality does not determine MLLM utility.}
    In the first example, the MLLM correctly identifies the author despite the
    unreadable reconstructed text. In the second, the reconstruction clearly
    preserves the book title, yet the MLLM produces an incorrect answer.}
    \label{fig:recon}
    \vspace{-3mm}
\end{figure}

\noindent\textbf{Hyperparameter Sensitivity.}
To assess hyperparameter sensitivity, we vary the neighborhood size $k$, and whitening regularization strength $\epsilon_w$ around their default values. As shown in Table~\ref{tab:hyperparameter_sensitivity}, macro Spearman and Pearson vary by at most \(0.021\) and \(0.023\), respectively, indicating that \momo{} does not rely on precise hyperparameter tuning.

 \noindent\textbf{Robustness of Test Set Size.}
We vary the number of paired examples from 500 to 5,000 while keeping the candidate encoder pool fixed. Figure~\ref{fig:robustness_test_size} shows that \momo{} maintains a high Spearman correlation throughout, with little gain from larger sets; several baselines are more sensitive to sample size. Thus, the smallest tested set already provides a useful ranking signal.

 \noindent\textbf{Robustness across Computational Budgets.}
As shown in Figure~\ref{fig:spearman_flops}, \momo{} maintains a Spearman correlation of approximately $0.84$ across a wide range of FLOPs, while competing methods achieve lower correlations despite generally higher costs.

\subsection{Reconstruction Quality as a Proxy for MLLM Performance}
\label{sec:recon}
Our candidate pool includes discrete visual tokenizers, which are commonly evaluated using
reconstruction-based metrics such as TokBench~\citep{wu2025tokbench}
metrics (T-ACC, T-NED, and F-Sim). We therefore examine whether these metrics predict their downstream MLLM performance.

 \noindent\textbf{Does reconstruction Reflect What the MLLM Can See?}
Reconstruction reflects what a dedicated decoder can recover from discrete
visual tokens, whereas downstream performance reflects whether the target LLM
can use them. To illustrate, we examine an MLLM built with TokLIP-L and
Qwen2.5-1.5B-Instruct on two examples from OCR-VQA~\citep{mishra2019ocrvqa}. As shown in Figure~\ref{fig:recon}, the
mismatch occurs in both directions: the MLLM correctly identifies ``Robert C.
Atkins'' despite severely distorted reconstructed text, but answers incorrectly
even when ``Banksy in New York'' is clearly reconstructed. Thus, accurate
reconstruction is neither necessary nor sufficient for an MLLM to extract
textual information from visual tokens.

 \noindent\textbf{Correlation with Downstream Performance.}
As shown in Table~\ref{tab:tokbench}, TokBench metrics do not consistently track downstream MLLM performance. Their Spearman correlations range from $+0.36$ on Qwen3 to $-0.97$ on SmolLM2, and Pearson correlations are likewise weak or negative. In contrast, RAVEL yields consistently positive correlations across all three LLMs, with Pearson correlations of $0.743$, $0.598$, and $0.524$, respectively. These results suggest that reconstruction quality is an unreliable criterion for selecting discrete visual tokenizers, while representation-level measures provide a more consistent signal of downstream utility.

%% file: sections/conclusion.tex
\section{Conclusion}
In this paper, we revisit whether vision encoder performance for a target LLM can be
estimated without training the corresponding MLLM. We introduce \momo{}, a
training-free method combining PCA whitening with patch-set matching over
paired image--text data. Across 210 MLLMs built from 70 vision encoders and
three language backbones, \momo{} achieves the strongest correlations with
downstream performance. Further analyses show that whitening mitigates
geometric bias, while patch-set matching preserves fine-grained relations,
making correspondence-aware local retrieval a simple and effective approach
to vision encoder selection.

%% file: sections/appendix.tex
\section{Algorithmic Description of \momo{}}

\label{app:momo_algorithm}
\begin{algorithm}[htbp]
\caption{\momo{} for vision encoder selection.}
\label{alg:momo}
\begin{algorithmic}[1]
\Require Encoders $\mathcal F=\{f_m\}_{m=1}^{M}$;
language model $g$;
paired data $\mathcal D=\{(I_i,T_i)\}_{i=1}^{n}$;
neighborhood size $k$;
stabilizer $\epsilon_w$
\Ensure Ranked encoders $\mathcal F_{\mathrm{ranked}}$

\Statex \textbf{Feature extraction}
\For{$i=1,\ldots,n$}
    \State $\displaystyle
    \mathbf t_i\gets
    \frac{1}{L_i}\sum_{\ell=1}^{L_i}
    \mathbf h_{i\ell}^{(-2)}$,
    where $\{\mathbf h_{i\ell}^{(-2)}\}_{\ell=1}^{L_i}$
    are the penultimate-layer text features
\EndFor
\For{$m=1,\ldots,M$}
    \For{$i=1,\ldots,n$}
        \State $\mathcal V_i^{(m)}
        \gets\{\mathbf p_{ia}^{(m)}\}_{a=1}^{P_i^{(m)}}$
        from $f_m(I_i)$, excluding special tokens
    \EndFor
\EndFor

\Statex \textbf{Independent stabilized whitening}
\State Compute $(\boldsymbol\mu^t,\mathbf U_+^t,\boldsymbol\Lambda_+^t)$
from $\{\mathbf t_i\}_{i=1}^{n}$, retaining positive eigenvalues
\For{$i=1,\ldots,n$}
    \State $\displaystyle
    \hat{\mathbf t}_i\gets
    \frac{
    (\boldsymbol\Lambda_+^t+\epsilon_w\mathbf I)^{-1/2}
    (\mathbf U_+^t)^\top(\mathbf t_i-\boldsymbol\mu^t)}
    {\left\|
    (\boldsymbol\Lambda_+^t+\epsilon_w\mathbf I)^{-1/2}
    (\mathbf U_+^t)^\top(\mathbf t_i-\boldsymbol\mu^t)
    \right\|_2}$
\EndFor

\For{$m=1,\ldots,M$}
    \State Compute
    $(\boldsymbol\mu^{v,m},\mathbf U_+^{v,m},
    \boldsymbol\Lambda_+^{v,m})$
    from all patches in $\bigcup_i\mathcal V_i^{(m)}$
    \ForAll{$\mathbf p_{ia}^{(m)}\in\mathcal V_i^{(m)}$}
        \State $\displaystyle
        \hat{\mathbf p}_{ia}^{(m)}
        \gets
        \frac{\mathbf q_{ia}^{(m)}}
        {\|\mathbf q_{ia}^{(m)}\|_2}$,
        where
        $\mathbf q_{ia}^{(m)}
        =
        (\boldsymbol\Lambda_+^{v,m}+\epsilon_w\mathbf I)^{-1/2}
        (\mathbf U_+^{v,m})^\top
        (\mathbf p_{ia}^{(m)}-\boldsymbol\mu^{v,m})$
    \EndFor
\EndFor

\Statex \textbf{Neighborhood construction}
\For{$i=1,\ldots,n$}
    \State $\displaystyle
    \mathcal N_i^t\gets
    \underset{j\neq i}{\arg \mathbf{topk}}\;
    \hat{\mathbf t}_i^\top\hat{\mathbf t}_j$
\EndFor

\For{$m=1,\ldots,M$}
    \ForAll{$1\leq i<j\leq n$}
        \State $\displaystyle
        s_{ij}^{v,m}\gets\frac{1}{2}
        \left[
        \frac{1}{P_i^{(m)}}\sum_a
        \max_b
        (\hat{\mathbf p}_{ia}^{(m)})^\top
        \hat{\mathbf p}_{jb}^{(m)}
        +
        \frac{1}{P_j^{(m)}}\sum_b
        \max_a
        (\hat{\mathbf p}_{jb}^{(m)})^\top
        \hat{\mathbf p}_{ia}^{(m)}
        \right]$
        \State $s_{ji}^{v,m}\gets s_{ij}^{v,m}$
    \EndFor
    \For{$i=1,\ldots,n$}
        \State $\displaystyle
        \mathcal N_i^{v,m}\gets
        \operatorname*{arg\,topk}_{j\neq i}\; s_{ij}^{v,m}$
    \EndFor
\EndFor

\Statex \textbf{Encoder scoring}
\For{$m=1,\ldots,M$}
    \State $\displaystyle
    \mathcal S_m\gets\frac{1}{nk}
    \sum_{i=1}^{n}
    \left|\mathcal N_i^{v,m}\cap\mathcal N_i^t\right|$
\EndFor
\State $\mathcal F_{\mathrm{ranked}}
\gets\{f_m\}_{m=1}^{M}$ ordered by decreasing $\mathcal S_m$
\State \Return $\mathcal F_{\mathrm{ranked}}$

\end{algorithmic}
\end{algorithm}

\begin{figure}[t]
    \centering
    \includegraphics[width=\linewidth]{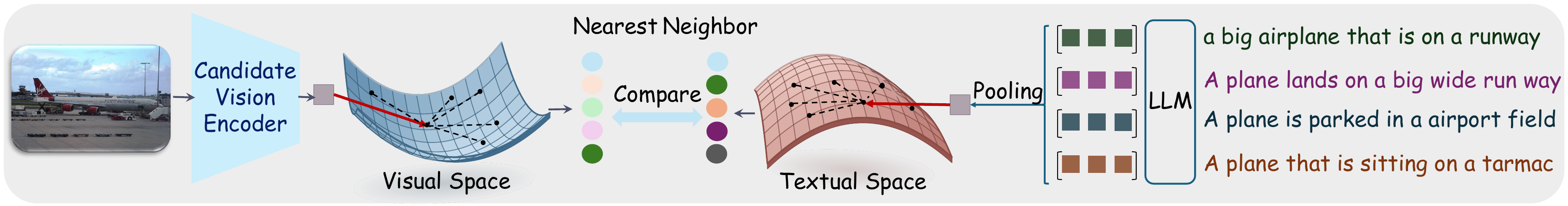}
    \caption{\textbf{Overview of \momo{}.}
    Given paired image--text samples, \momo{} applies stabilized PCA whitening,
    constructs visual and textual neighborhoods using patch-level and global
    similarity, respectively, and measures their agreement through the known
    image--text correspondences.}
    \label{fig:method}
\end{figure}

\section{Overview of \ravel}
\label{APP:overview}
Figure~\ref{fig:method} summarizes the overall pipeline. Given paired
image--text samples, \momo{} extracts visual patch features from each candidate
encoder and text features from the target LLM, then whitens each representation
space independently. It constructs textual neighborhoods using cosine
similarity and visual neighborhoods using symmetric patch-level matching.
Each encoder is scored by the average overlap between these neighborhoods under
the known image--text correspondences. The procedure uses only frozen features
and requires no downstream labels.

\section{Candidate Vision Encoders}
\label{app:model}
Our candidate pool contains 70 vision encoder configurations spanning
continuous encoders and discrete or hybrid visual tokenizers.
Table~\ref{tab:vision_encoders} lists all evaluated architecture, checkpoint,
and resolution variants, treating each resolution of an architecture as a
separate configuration.

\begin{table}[t]
    \centering
    \caption{\textbf{Complete vision encoder pool.}
    All 70 evaluated architecture, checkpoint, and input-resolution configurations
    are listed and grouped by model family.}
    \label{tab:vision_encoders}

    \scriptsize
    \setlength{\tabcolsep}{2pt}
    \renewcommand{\arraystretch}{1.04}

    \begin{tabularx}{\columnwidth}{
        @{}
        >{\raggedright\arraybackslash}p{0.20\columnwidth}
        >{\raggedright\arraybackslash}X
        >{\raggedright\arraybackslash}p{0.20\columnwidth}
        @{}
    }
        \toprule
        \textbf{Family (\#)}
        & \textbf{Architectures and input resolutions}
        & \textbf{Reference} \\
        \midrule

        \multicolumn{3}{@{}l}{
            \textit{Language-supervised continuous encoders}
        } \\

        OpenAI CLIP (1)
            & ViT-L/14 (224)
            & \citealp{radford2021clip} \\

        MetaCLIP (9)
            & ViT-B/16 (400M, 224; 2.5B, 224);
              ViT-B/32 (400M, 224; 2.5B, 224);
              ViT-L/14 (400M, 224; 2.5B, 224);
              ViT-H/14 (2.5B, 224; v1.2, 224);
              ViT-G/14 (2.5B, 224)
            & \citealp{xu2024metaclip} \\

        MetaCLIP~2 (15)
            & ViT-B/16 (224, 384);
              ViT-B/32 (224, 384; mT5, 224);
              ViT-G/14 (224, 378);
              ViT-H/14 (378);
              ViT-L/14 (224);
              ViT-M/16 (224, 384; mT5, 224);
              ViT-S/16 (224, 384; mT5, 224)
            & \citealp{chuang2025metaclip2} \\

        SigLIP2 (15)
            & So400m/14 (224, 384);
              So400m/16 (256, 384, 512);
              ViT-B/16 (224, 256, 384, 512);
              ViT-B/32 (256);
              ViT-G/16 (256, 384);
              ViT-L/16 (256, 384, 512)
            & \citealp{tschannen2025siglip2} \\

        Perception Encoder (3)
            & PE-Core-B/16 (224);
              PE-Core-G/14 (448);
              PE-Lang-L/14 (448)
            & \citealp{bolya2025perception} \\

        \midrule
        \multicolumn{3}{@{}l}{
            \textit{Image-only self-supervised continuous encoders}
        } \\

        DINO (4)
            & ViT-S/8; ViT-S/16; ViT-B/8; ViT-B/16
            & \citealp{caron2021dino} \\

        DINOv2 (4)
            & ViT-S/14; ViT-B/14; ViT-L/14; ViT-G/14
            & \citealp{oquab2023dinov2} \\

        DINOv3 (1)
            & ViT-L/16
            & \citealp{simeoni2025dinov3} \\

        I-JEPA (1)
            & ViT-H/14
            & \citealp{assran2023ijepa} \\

        Web-SSL DINO (1)
            & 1B (224)
            & \citealp{fan2025webssl} \\

        \midrule
        \multicolumn{3}{@{}l}{
            \textit{Reconstruction-oriented continuous encoders}
        } \\

        Web-SSL MAE (3)
            & 300M (224); 1B (224); 3B (224)
            & \citealp{fan2025webssl} \\

        Pixio (3)
            & ViT-B/16; ViT-L/16; ViT-H/16
            & \citealp{yang2026pixio} \\

        RAEv2 (1)
            & DINOv3-L
            & \citealp{singh2026raev2} \\

        \midrule
        \multicolumn{3}{@{}l}{
            \textit{Distillation-based continuous encoders}
        } \\

        EUPE (4)
            & ViT-T; ViT-S; ViT-B; ConvNeXt-B
            & \citealp{zhu2026eupe} \\

        \midrule
        \multicolumn{3}{@{}l}{
            \textit{Discrete and hybrid visual tokenizers}
        } \\

        TokLIP (2)
            & TokLIP-S (256); TokLIP-L (384)
            & \citealp{lin2025toklip} \\

        UniTok (1)
            & UniTok (256)
            & \citealp{ma2025unitok} \\

        VILA-U (1)
            & VILA-U (256)
            & \citealp{wu2025vilau} \\

        UniAR (1)
            & UniAR-BSQ
            & \citealp{peng2026unified} \\

        \bottomrule
    \end{tabularx}
\end{table}

\section{Detailed Downstream MLLM Results}
\label{app:detailed_mllm_results}

Table~\ref{tab:detailed_mllm_all} reports the complete downstream results of 70 vision encoders paired with three language backbones, yielding 210 MLLMs. Each row corresponds to one encoder--backbone pair and includes performance on eleven benchmarks spanning visual question answering, reasoning, document and chart understanding, object hallucination, and image captioning. \textsc{Avg}, the unweighted mean across these benchmarks, defines the ground-truth performance and global ranking.

Language-supervised encoders, particularly models from the SigLIP2 and MetaCLIP families, consistently occupy the top ranks. However, the optimal choice remains strongly dependent on the language backbone. Specifically, the best-performing configurations are SigLIP2 ViT-G/16 at 384~pixels for Qwen3 (\(54.56\)), SigLIP2 So400m/16 at 512~pixels for Qwen2.5 (\(54.97\)), and MetaCLIP~2 ViT-G/14 at 378~pixels for SmolLM2 (\(45.93\)). These differences confirm that no single vision encoder is uniformly optimal across language backbones.

\begingroup
\scriptsize
\setlength{\tabcolsep}{2pt}
\renewcommand{\arraystretch}{0.92}
\setlength{\LTleft}{0pt}
\setlength{\LTright}{0pt}

\newcolumntype{Z}{>{\raggedright\arraybackslash}l}
\newcolumntype{R}{>{\raggedleft\arraybackslash}X}

\begin{xltabular}{\linewidth}{@{}p{0.52cm}cZ*{12}{R}@{}}
\caption{
Detailed downstream MLLM results for all 70 vision encoders and three language
backbones. The 210 encoder--language-backbone pairs are sorted globally by
\textsc{Avg} in descending order. The labels \textsc{Q3}, \textsc{Q2.5}, and
\textsc{Smol} denote Qwen3-1.7B, Qwen2.5-1.5B-Instruct, and
SmolLM2-1.7B-Instruct, respectively. The background colors identify the
language backbone.
}
\label{tab:detailed_mllm_all}
\\
\toprule
\rowcolor{HeaderTeal}
\cellcolor{HeaderTeal}\textcolor{white}{\textbf{LLM}} &
\cellcolor{HeaderTeal}\textcolor{white}{\textbf{Rank}} &
\cellcolor{HeaderTeal}\textcolor{white}{\textbf{Vision Encoder}} &
\cellcolor{HeaderTeal}\textcolor{white}{\shortstack{\tiny\textbf{MM}\\\tiny\textbf{MU}}} &
\cellcolor{HeaderTeal}\textcolor{white}{\tiny\textbf{MMB}} &
\cellcolor{HeaderTeal}\textcolor{white}{\shortstack{\tiny\textbf{VQA}\\\tiny\textbf{v2}}} &
\cellcolor{HeaderTeal}\textcolor{white}{\tiny\textbf{SQA}} &
\cellcolor{HeaderTeal}\textcolor{white}{\tiny\textbf{Chart}} &
\cellcolor{HeaderTeal}\textcolor{white}{\tiny\textbf{Doc}} &
\cellcolor{HeaderTeal}\textcolor{white}{\tiny\textbf{Text}} &
\cellcolor{HeaderTeal}\textcolor{white}{\tiny\textbf{POPE}} &
\cellcolor{HeaderTeal}\textcolor{white}{\tiny\textbf{GQA}} &
\cellcolor{HeaderTeal}\textcolor{white}{\tiny\textbf{COCO}} &
\cellcolor{HeaderTeal}\textcolor{white}{\tiny\textbf{Flickr}} &
\cellcolor{HeaderTeal}\textcolor{white}{\tiny\textbf{Avg}} \\
\midrule
\endfirsthead

\multicolumn{15}{c}{\tablename\ \thetable{} -- continued from previous page}\\
\toprule
\rowcolor{HeaderTeal}
\cellcolor{HeaderTeal}\textcolor{white}{\textbf{LLM}} &
\cellcolor{HeaderTeal}\textcolor{white}{\textbf{Rank}} &
\cellcolor{HeaderTeal}\textcolor{white}{\textbf{Vision Encoder}} &
\cellcolor{HeaderTeal}\textcolor{white}{\shortstack{\tiny\textbf{MM}\\\tiny\textbf{MU}}} &
\cellcolor{HeaderTeal}\textcolor{white}{\tiny\textbf{MMB}} &
\cellcolor{HeaderTeal}\textcolor{white}{\shortstack{\tiny\textbf{VQA}\\\tiny\textbf{v2}}} &
\cellcolor{HeaderTeal}\textcolor{white}{\tiny\textbf{SQA}} &
\cellcolor{HeaderTeal}\textcolor{white}{\tiny\textbf{Chart}} &
\cellcolor{HeaderTeal}\textcolor{white}{\tiny\textbf{Doc}} &
\cellcolor{HeaderTeal}\textcolor{white}{\tiny\textbf{Text}} &
\cellcolor{HeaderTeal}\textcolor{white}{\tiny\textbf{POPE}} &
\cellcolor{HeaderTeal}\textcolor{white}{\tiny\textbf{GQA}} &
\cellcolor{HeaderTeal}\textcolor{white}{\tiny\textbf{COCO}} &
\cellcolor{HeaderTeal}\textcolor{white}{\tiny\textbf{Flickr}} &
\cellcolor{HeaderTeal}\textcolor{white}{\tiny\textbf{Avg}} \\
\midrule
\endhead

\midrule
\multicolumn{15}{r}{Continued on next page}\\
\endfoot

\bottomrule
\endlastfoot

\rowcolor{Qwen25Lavender!14}
\cellcolor{Qwen25Lavender!50}\textbf{\tiny Q2.5} & 1 & SigLIP2 So400m/16 (512) & 36.10 & 64.86 & 67.52 & 65.50 & 14.10 & \textbf{26.22} & 47.79 & 84.91 & \textbf{55.49} & 85.14 & 57.01 & \textbf{54.97} \\
\rowcolor{Qwen3Mint!12}
\cellcolor{Qwen3Mint!45}\textbf{\tiny Q3} & 2 & SigLIP2 ViT-G/16 (384) & 32.40 & 61.39 & \textbf{67.72} & 68.40 & 14.40 & 25.22 & \textbf{48.88} & \textbf{91.82} & \textbf{54.47} & \textbf{81.97} & 53.53 & \textbf{54.56} \\
\rowcolor{Qwen25Lavender!14}
\cellcolor{Qwen25Lavender!50}\textbf{\tiny Q2.5} & 3 & SigLIP2 So400m/14 (384) & 35.50 & 66.02 & 67.89 & 65.90 & 14.10 & 24.59 & 46.61 & 89.98 & 52.64 & 82.34 & 53.82 & 54.49 \\
\rowcolor{Qwen25Lavender!14}
\cellcolor{Qwen25Lavender!50}\textbf{\tiny Q2.5} & 4 & SigLIP2 ViT-L/16 (512) & 35.00 & 64.09 & 66.63 & \textbf{67.10} & 13.20 & 25.07 & 45.45 & 91.39 & 55.08 & 80.59 & 55.53 & 54.47 \\
\rowcolor{Qwen25Lavender!14}
\cellcolor{Qwen25Lavender!50}\textbf{\tiny Q2.5} & 5 & SigLIP2 ViT-G/16 (384) & 34.10 & \textbf{70.27} & 68.07 & 66.80 & \textbf{15.10} & 24.62 & \textbf{50.33} & 84.65 & 53.56 & 77.92 & 51.37 & 54.25 \\
\rowcolor{Qwen3Mint!12}
\cellcolor{Qwen3Mint!45}\textbf{\tiny Q3} & 6 & SigLIP2 So400m/14 (384) & 33.30 & 62.93 & 66.39 & \textbf{69.20} & 14.20 & 25.62 & 47.33 & 91.61 & 52.03 & 78.40 & \textbf{53.60} & 54.06 \\
\rowcolor{Qwen25Lavender!14}
\cellcolor{Qwen25Lavender!50}\textbf{\tiny Q2.5} & 7 & SigLIP2 ViT-G/16 (256) & 34.60 & 66.80 & \textbf{68.08} & 66.80 & 12.40 & 15.92 & 39.51 & 91.21 & 53.86 & \textbf{85.89} & \textbf{57.45} & 53.87 \\
\rowcolor{Qwen25Lavender!14}
\cellcolor{Qwen25Lavender!50}\textbf{\tiny Q2.5} & 8 & SigLIP2 So400m/16 (384) & 37.10 & 64.48 & 66.25 & 64.00 & 12.90 & 22.49 & 45.26 & 91.58 & 53.25 & 72.46 & 54.53 & 53.12 \\
\rowcolor{Qwen25Lavender!14}
\cellcolor{Qwen25Lavender!50}\textbf{\tiny Q2.5} & 9 & SigLIP2 So400m/14 (224) & 35.70 & 66.02 & 66.87 & 66.40 & 12.70 & 15.73 & 37.55 & 89.41 & 52.24 & 84.10 & 56.51 & 53.02 \\
\rowcolor{Qwen25Lavender!14}
\cellcolor{Qwen25Lavender!50}\textbf{\tiny Q2.5} & 10 & SigLIP2 ViT-L/16 (384) & 35.20 & 64.09 & 64.21 & 66.00 & 13.90 & 18.89 & 41.45 & 89.73 & 52.64 & 84.15 & 52.84 & 53.01 \\
\rowcolor{Qwen3Mint!12}
\cellcolor{Qwen3Mint!45}\textbf{\tiny Q3} & 11 & SigLIP2 ViT-L/16 (384) & 32.70 & 63.71 & 65.14 & 66.30 & \textbf{14.70} & 21.45 & 45.17 & 90.14 & 53.15 & 78.08 & 47.60 & 52.56 \\
\rowcolor{Qwen25Lavender!14}
\cellcolor{Qwen25Lavender!50}\textbf{\tiny Q2.5} & 12 & MetaCLIP 2 ViT-G/14 (378) & \textbf{38.10} & 66.41 & 65.04 & 65.00 & 13.60 & 20.62 & 39.14 & 88.82 & 52.95 & 76.55 & 51.97 & 52.56 \\
\rowcolor{Qwen3Mint!12}
\cellcolor{Qwen3Mint!45}\textbf{\tiny Q3} & 13 & SigLIP2 So400m/16 (512) & 32.80 & 62.16 & 64.16 & 68.20 & 14.00 & \textbf{26.22} & 47.02 & 87.61 & 50.20 & 75.77 & 49.44 & 52.51 \\
\rowcolor{Qwen25Lavender!14}
\cellcolor{Qwen25Lavender!50}\textbf{\tiny Q2.5} & 14 & PE-Core-G/14 (448) & 34.60 & 66.02 & 62.34 & 64.30 & 14.00 & 21.12 & 39.97 & 89.55 & 49.49 & 77.00 & 52.32 & 51.88 \\
\rowcolor{Qwen3Mint!12}
\cellcolor{Qwen3Mint!45}\textbf{\tiny Q3} & 15 & SigLIP2 So400m/16 (384) & 33.20 & 62.55 & 64.76 & 67.50 & 13.10 & 23.48 & 47.77 & 86.59 & 54.07 & 70.64 & 45.61 & 51.75 \\
\rowcolor{Qwen25Lavender!14}
\cellcolor{Qwen25Lavender!50}\textbf{\tiny Q2.5} & 16 & SigLIP2 So400m/16 (256) & 36.70 & 64.09 & 66.69 & 62.50 & 12.40 & 17.54 & 37.83 & \textbf{91.60} & 52.54 & 74.17 & 49.39 & 51.40 \\
\rowcolor{Qwen25Lavender!14}
\cellcolor{Qwen25Lavender!50}\textbf{\tiny Q2.5} & 17 & MetaCLIP ViT-H/14 (v1.2, 224) & 37.30 & 66.41 & 64.56 & 65.00 & 11.10 & 14.72 & 34.05 & 83.59 & 52.95 & 82.66 & 50.42 & 51.16 \\
\rowcolor{Qwen3Mint!12}
\cellcolor{Qwen3Mint!45}\textbf{\tiny Q3} & 18 & SigLIP2 ViT-G/16 (256) & 33.50 & \textbf{67.18} & 64.62 & 65.30 & 11.60 & 16.59 & 39.18 & 87.88 & 52.74 & 76.16 & 46.45 & 51.02 \\
\rowcolor{Qwen25Lavender!14}
\cellcolor{Qwen25Lavender!50}\textbf{\tiny Q2.5} & 19 & MetaCLIP 2 ViT-H/14 (378) & 35.00 & 65.64 & 62.40 & 63.60 & 12.80 & 21.13 & 37.22 & 82.60 & 51.22 & 77.40 & 50.73 & 50.89 \\
\rowcolor{Qwen25Lavender!14}
\cellcolor{Qwen25Lavender!50}\textbf{\tiny Q2.5} & 20 & MetaCLIP 2 ViT-G/14 (224) & 36.20 & 64.86 & 63.96 & 63.50 & 11.50 & 15.44 & 34.35 & 89.54 & 51.93 & 76.75 & 50.45 & 50.77 \\
\rowcolor{Qwen25Lavender!14}
\cellcolor{Qwen25Lavender!50}\textbf{\tiny Q2.5} & 21 & SigLIP2 ViT-L/16 (256) & 34.50 & 63.71 & 64.12 & 65.40 & 12.30 & 15.44 & 38.19 & 89.32 & 51.32 & 75.60 & 48.44 & 50.76 \\
\rowcolor{Qwen25Lavender!14}
\cellcolor{Qwen25Lavender!50}\textbf{\tiny Q2.5} & 22 & SigLIP2 ViT-B/16 (384) & 33.50 & 64.86 & 63.90 & 64.80 & 12.30 & 17.65 & 34.25 & 89.26 & 51.32 & 75.79 & 47.91 & 50.50 \\
\rowcolor{Qwen25Lavender!14}
\cellcolor{Qwen25Lavender!50}\textbf{\tiny Q2.5} & 23 & SigLIP2 ViT-B/16 (512) & 34.70 & 63.32 & 62.78 & 64.40 & 13.00 & 20.67 & 35.90 & 88.67 & 51.22 & 72.31 & 44.84 & 50.16 \\
\rowcolor{Qwen25Lavender!14}
\cellcolor{Qwen25Lavender!50}\textbf{\tiny Q2.5} & 24 & MetaCLIP ViT-G/14 (2.5B, 224) & 35.00 & 69.11 & 63.56 & 65.40 & 11.50 & 14.84 & 31.90 & 88.52 & 50.71 & 69.57 & 48.40 & 49.86 \\
\rowcolor{Qwen25Lavender!14}
\cellcolor{Qwen25Lavender!50}\textbf{\tiny Q2.5} & 25 & OpenAI CLIP ViT-L/14 (224) & 33.60 & 66.41 & 61.17 & 64.60 & 11.80 & 14.17 & 32.56 & 89.18 & 51.32 & 74.11 & 48.42 & 49.76 \\
\rowcolor{Qwen25Lavender!14}
\cellcolor{Qwen25Lavender!50}\textbf{\tiny Q2.5} & 26 & MetaCLIP 2 ViT-L/14 (224) & 34.50 & 67.18 & 60.84 & 63.20 & 12.50 & 14.75 & 32.11 & 88.67 & 51.32 & 71.15 & 47.88 & 49.46 \\
\rowcolor{Qwen3Mint!12}
\cellcolor{Qwen3Mint!45}\textbf{\tiny Q3} & 27 & SigLIP2 So400m/16 (256) & 32.00 & 59.85 & 64.64 & 67.70 & 12.30 & 16.84 & 39.13 & 90.42 & 51.93 & 64.35 & 41.16 & 49.12 \\
\rowcolor{Qwen3Mint!12}
\cellcolor{Qwen3Mint!45}\textbf{\tiny Q3} & 28 & SigLIP2 So400m/14 (224) & 33.20 & 61.78 & 64.70 & 65.80 & 12.90 & 14.88 & 36.69 & 90.04 & 53.25 & 65.91 & 41.12 & 49.12 \\
\rowcolor{Qwen3Mint!12}
\cellcolor{Qwen3Mint!45}\textbf{\tiny Q3} & 29 & SigLIP2 ViT-L/16 (256) & 33.30 & 61.00 & 62.45 & 66.90 & 11.30 & 14.39 & 35.96 & 83.08 & 52.54 & 74.30 & 44.81 & 49.09 \\
\rowcolor{Qwen25Lavender!14}
\cellcolor{Qwen25Lavender!50}\textbf{\tiny Q2.5} & 30 & MetaCLIP ViT-H/14 (2.5B, 224) & 33.10 & 62.55 & 61.08 & 64.60 & 10.80 & 15.26 & 28.23 & 88.65 & 51.22 & 74.23 & 48.40 & 48.92 \\
\rowcolor{Qwen3Mint!12}
\cellcolor{Qwen3Mint!45}\textbf{\tiny Q3} & 31 & MetaCLIP 2 ViT-G/14 (224) & 32.70 & 63.71 & 62.16 & 64.30 & 11.70 & 14.43 & 32.55 & 85.93 & 51.22 & 73.08 & 45.49 & 48.84 \\
\rowcolor{Qwen25Lavender!14}
\cellcolor{Qwen25Lavender!50}\textbf{\tiny Q2.5} & 32 & MetaCLIP ViT-L/14 (2.5B, 224) & 36.90 & 66.80 & 61.34 & 63.50 & 12.20 & 13.07 & 29.38 & 81.04 & 51.73 & 72.73 & 47.87 & 48.78 \\
\rowcolor{Qwen3Mint!12}
\cellcolor{Qwen3Mint!45}\textbf{\tiny Q3} & 33 & MetaCLIP ViT-H/14 (v1.2, 224) & 34.10 & 61.78 & 63.81 & 63.90 & 10.90 & 13.80 & 31.48 & 87.87 & 49.70 & 72.17 & 45.88 & 48.67 \\
\rowcolor{Qwen3Mint!12}
\cellcolor{Qwen3Mint!45}\textbf{\tiny Q3} & 34 & PE-Core-G/14 (448) & 34.10 & 62.93 & 59.88 & 66.10 & 14.20 & 16.32 & 34.12 & 79.83 & 48.58 & 72.55 & 45.76 & 48.58 \\
\rowcolor{Qwen3Mint!12}
\cellcolor{Qwen3Mint!45}\textbf{\tiny Q3} & 35 & SigLIP2 ViT-B/16 (512) & 32.60 & 60.23 & 60.41 & 61.70 & 13.70 & 18.74 & 38.13 & 86.31 & 51.52 & 67.14 & 43.12 & 48.51 \\
\rowcolor{Qwen3Mint!12}
\cellcolor{Qwen3Mint!45}\textbf{\tiny Q3} & 36 & OpenAI CLIP ViT-L/14 (224) & 33.70 & 62.55 & 61.16 & 64.60 & 12.40 & 14.73 & 32.85 & 85.21 & 49.59 & 68.61 & 40.71 & 47.83 \\
\rowcolor{Qwen25Lavender!14}
\cellcolor{Qwen25Lavender!50}\textbf{\tiny Q2.5} & 37 & SigLIP2 ViT-B/16 (224) & 32.70 & 64.86 & 63.27 & 65.10 & 11.20 & 11.83 & 26.96 & 78.87 & 53.05 & 73.94 & 44.37 & 47.83 \\
\rowcolor{Qwen3Mint!12}
\cellcolor{Qwen3Mint!45}\textbf{\tiny Q3} & 38 & MetaCLIP ViT-G/14 (2.5B, 224) & 34.60 & 61.39 & 61.24 & 64.30 & 12.40 & 13.85 & 31.20 & 84.21 & 50.10 & 68.09 & 43.33 & 47.70 \\
\rowcolor{Qwen25Lavender!14}
\cellcolor{Qwen25Lavender!50}\textbf{\tiny Q2.5} & 39 & SigLIP2 ViT-B/16 (256) & 33.70 & 64.86 & 62.53 & 64.70 & 12.00 & 11.89 & 29.16 & 82.00 & 51.63 & 68.42 & 42.89 & 47.62 \\
\rowcolor{Qwen3Mint!12}
\cellcolor{Qwen3Mint!45}\textbf{\tiny Q3} & 40 & MetaCLIP 2 ViT-G/14 (378) & 33.10 & 60.62 & 59.70 & 63.20 & 11.90 & 15.59 & 31.83 & 84.43 & 48.98 & 67.13 & 41.36 & 47.08 \\
\rowcolor{Qwen25Lavender!14}
\cellcolor{Qwen25Lavender!50}\textbf{\tiny Q2.5} & 41 & MetaCLIP ViT-L/14 (400M, 224) & 33.80 & 62.16 & 61.16 & 62.60 & 11.40 & 13.72 & 28.16 & 79.71 & 50.30 & 70.94 & 43.90 & 47.08 \\
\rowcolor{Qwen3Mint!12}
\cellcolor{Qwen3Mint!45}\textbf{\tiny Q3} & 42 & SigLIP2 ViT-B/16 (256) & 33.80 & 61.39 & 61.12 & 66.90 & 11.10 & 13.77 & 28.42 & 89.73 & 50.00 & 63.77 & 36.01 & 46.91 \\
\rowcolor{Qwen25Lavender!14}
\cellcolor{Qwen25Lavender!50}\textbf{\tiny Q2.5} & 43 & MetaCLIP 2 ViT-B/16 (384) & 33.30 & 61.00 & 59.22 & 63.60 & 11.70 & 14.31 & 30.93 & 82.39 & 51.02 & 67.51 & 40.40 & 46.85 \\
\rowcolor{Qwen3Mint!12}
\cellcolor{Qwen3Mint!45}\textbf{\tiny Q3} & 44 & SigLIP2 ViT-B/16 (224) & 31.50 & 58.30 & 59.71 & 65.20 & 12.00 & 13.38 & 28.24 & 89.31 & 51.83 & 64.23 & 38.95 & 46.60 \\
\rowcolor{Qwen3Mint!12}
\cellcolor{Qwen3Mint!45}\textbf{\tiny Q3} & 45 & MetaCLIP ViT-H/14 (2.5B, 224) & \textbf{35.40} & 58.69 & 59.53 & 63.60 & 12.00 & 13.13 & 28.98 & 86.79 & 50.30 & 64.02 & 37.94 & 46.40 \\
\rowcolor{Qwen3Mint!12}
\cellcolor{Qwen3Mint!45}\textbf{\tiny Q3} & 46 & MetaCLIP ViT-L/14 (2.5B, 224) & 34.10 & 60.23 & 60.09 & 65.80 & 12.60 & 11.37 & 29.32 & 85.47 & 47.05 & 63.04 & 37.85 & 46.08 \\
\rowcolor{Qwen25Lavender!14}
\cellcolor{Qwen25Lavender!50}\textbf{\tiny Q2.5} & 47 & Pixio ViT-L/16 & 33.80 & 57.53 & 58.32 & 62.60 & 11.30 & 8.26 & 11.23 & 90.08 & 53.96 & 74.56 & 43.76 & 45.95 \\
\rowcolor{SmolPeach!16}
\cellcolor{SmolPeach!55}\textbf{\tiny Smol} & 48 & MetaCLIP 2 ViT-G/14 (378) & 31.10 & 57.53 & \textbf{61.97} & \textbf{60.40} & \textbf{13.70} & 17.39 & 38.08 & 84.41 & 48.88 & 56.77 & 35.04 & \textbf{45.93} \\
\rowcolor{Qwen25Lavender!14}
\cellcolor{Qwen25Lavender!50}\textbf{\tiny Q2.5} & 49 & PE-Core-B/16 (224) & 34.30 & 60.62 & 60.92 & 61.70 & 11.60 & 10.91 & 23.63 & 83.13 & 51.52 & 65.44 & 41.25 & 45.91 \\
\rowcolor{Qwen3Mint!12}
\cellcolor{Qwen3Mint!45}\textbf{\tiny Q3} & 50 & MetaCLIP ViT-L/14 (400M, 224) & 32.70 & 57.53 & 58.43 & 65.60 & 12.00 & 11.35 & 27.07 & 85.81 & 48.58 & 67.63 & 38.04 & 45.89 \\
\rowcolor{SmolPeach!16}
\cellcolor{SmolPeach!55}\textbf{\tiny Smol} & 51 & SigLIP2 ViT-L/16 (384) & 27.40 & 57.14 & 59.62 & 55.10 & 11.80 & 16.92 & 39.32 & 80.72 & 49.80 & 61.74 & \textbf{44.85} & 45.86 \\
\rowcolor{Qwen25Lavender!14}
\cellcolor{Qwen25Lavender!50}\textbf{\tiny Q2.5} & 52 & PE-Lang-L/14 (448) & 32.50 & 57.53 & 59.69 & 61.00 & 12.70 & 9.80 & 22.32 & 83.59 & 50.91 & 68.55 & 43.26 & 45.62 \\
\rowcolor{SmolPeach!16}
\cellcolor{SmolPeach!55}\textbf{\tiny Smol} & 53 & SigLIP2 So400m/14 (384) & 28.70 & 56.76 & 59.95 & 51.00 & 13.20 & 18.78 & 42.03 & \textbf{84.84} & \textbf{50.61} & 55.26 & 40.12 & 45.57 \\
\rowcolor{Qwen25Lavender!14}
\cellcolor{Qwen25Lavender!50}\textbf{\tiny Q2.5} & 54 & MetaCLIP 2 ViT-B/16 (224) & 35.30 & 59.85 & 59.43 & 60.50 & 11.10 & 11.06 & 23.81 & 85.85 & 47.56 & 66.64 & 40.07 & 45.56 \\
\rowcolor{Qwen3Mint!12}
\cellcolor{Qwen3Mint!45}\textbf{\tiny Q3} & 55 & PE-Core-B/16 (224) & 33.10 & 58.30 & 58.47 & 64.80 & 11.50 & 11.47 & 24.93 & 87.57 & 49.29 & 63.94 & 37.20 & 45.51 \\
\rowcolor{Qwen25Lavender!14}
\cellcolor{Qwen25Lavender!50}\textbf{\tiny Q2.5} & 56 & SigLIP2 ViT-B/32 (256) & 34.70 & 62.93 & 59.38 & 65.40 & 11.30 & 9.03 & 17.92 & 83.46 & 49.19 & 65.03 & 39.16 & 45.23 \\
\rowcolor{Qwen25Lavender!14}
\cellcolor{Qwen25Lavender!50}\textbf{\tiny Q2.5} & 57 & MetaCLIP 2 ViT-M/16 (384) & 35.10 & 58.69 & 57.77 & 61.70 & 11.40 & 14.53 & 26.82 & 84.11 & 47.97 & 61.26 & 33.30 & 44.79 \\
\rowcolor{Qwen25Lavender!14}
\cellcolor{Qwen25Lavender!50}\textbf{\tiny Q2.5} & 58 & MetaCLIP ViT-B/16 (400M, 224) & 34.60 & 57.14 & 58.20 & 63.90 & 11.10 & 11.69 & 22.61 & 86.83 & 49.29 & 62.82 & 34.45 & 44.78 \\
\rowcolor{Qwen25Lavender!14}
\cellcolor{Qwen25Lavender!50}\textbf{\tiny Q2.5} & 59 & MetaCLIP ViT-B/16 (2.5B, 224) & 33.10 & 62.55 & 59.85 & 62.80 & 10.80 & 11.84 & 25.07 & 78.96 & 50.30 & 61.13 & 35.75 & 44.74 \\
\rowcolor{Qwen25Lavender!14}
\cellcolor{Qwen25Lavender!50}\textbf{\tiny Q2.5} & 60 & DINOv2 ViT-G/14 & 34.20 & 55.98 & 59.08 & 60.90 & 10.90 & 7.81 & 10.74 & 88.81 & 50.61 & 68.50 & 41.08 & 44.42 \\
\rowcolor{Qwen3Mint!12}
\cellcolor{Qwen3Mint!45}\textbf{\tiny Q3} & 61 & MetaCLIP 2 ViT-B/16 (384) & 32.70 & 55.98 & 58.06 & 62.80 & 11.80 & 14.55 & 28.86 & 77.97 & 47.26 & 63.44 & 33.80 & 44.29 \\
\rowcolor{Qwen25Lavender!14}
\cellcolor{Qwen25Lavender!50}\textbf{\tiny Q2.5} & 62 & Web-SSL DINO 1B (224) & 32.10 & 58.30 & 58.17 & 62.30 & 10.50 & 8.26 & 11.92 & 81.52 & 49.19 & 73.07 & 41.55 & 44.26 \\
\rowcolor{SmolPeach!16}
\cellcolor{SmolPeach!55}\textbf{\tiny Smol} & 63 & SigLIP2 So400m/16 (384) & 26.90 & 57.53 & 60.78 & 54.30 & 13.30 & 17.40 & 40.36 & 66.36 & 46.95 & 63.33 & 37.68 & 44.08 \\
\rowcolor{Qwen3Mint!12}
\cellcolor{Qwen3Mint!45}\textbf{\tiny Q3} & 64 & MetaCLIP ViT-B/16 (400M, 224) & 34.00 & 57.53 & 57.91 & 65.10 & 12.00 & 11.21 & 23.76 & 85.13 & 48.78 & 58.09 & 31.04 & 44.05 \\
\rowcolor{Qwen25Lavender!14}
\cellcolor{Qwen25Lavender!50}\textbf{\tiny Q2.5} & 65 & EUPE ConvNeXt-B & 34.00 & 59.46 & 56.96 & 62.30 & 10.50 & 8.75 & 16.76 & 80.17 & 49.29 & 66.26 & 40.01 & 44.04 \\
\rowcolor{Qwen25Lavender!14}
\cellcolor{Qwen25Lavender!50}\textbf{\tiny Q2.5} & 66 & TokLIP-L (384) & 35.30 & 59.07 & 57.26 & 64.20 & 10.80 & 8.87 & 12.26 & 72.99 & 50.00 & 71.19 & 41.93 & 43.99 \\
\rowcolor{Qwen3Mint!12}
\cellcolor{Qwen3Mint!45}\textbf{\tiny Q3} & 67 & MetaCLIP ViT-B/16 (2.5B, 224) & 33.20 & 54.83 & 55.86 & 62.80 & 11.50 & 10.71 & 23.13 & 82.14 & 49.39 & 63.61 & 35.68 & 43.90 \\
\rowcolor{SmolPeach!16}
\cellcolor{SmolPeach!55}\textbf{\tiny Smol} & 68 & SigLIP2 ViT-G/16 (256) & 29.50 & 58.69 & 57.52 & 56.00 & 12.00 & 14.52 & 37.87 & 65.31 & 46.04 & 60.52 & 43.33 & 43.75 \\
\rowcolor{Qwen3Mint!12}
\cellcolor{Qwen3Mint!45}\textbf{\tiny Q3} & 69 & MetaCLIP 2 ViT-B/16 (224) & 33.80 & 57.14 & 57.02 & 62.30 & 10.60 & 11.42 & 26.23 & 83.47 & 47.46 & 58.59 & 32.08 & 43.65 \\
\rowcolor{Qwen3Mint!12}
\cellcolor{Qwen3Mint!45}\textbf{\tiny Q3} & 70 & Web-SSL DINO 1B (224) & 32.70 & 56.37 & 57.60 & 61.90 & 10.40 & 7.38 & 12.19 & 88.19 & 49.39 & 68.14 & 35.87 & 43.65 \\
\rowcolor{Qwen3Mint!12}
\cellcolor{Qwen3Mint!45}\textbf{\tiny Q3} & 71 & MetaCLIP 2 ViT-B/32 (384) & 32.90 & 55.21 & 55.04 & 64.60 & 11.20 & 10.44 & 20.35 & 84.27 & 47.56 & 62.28 & 33.24 & 43.37 \\
\rowcolor{Qwen25Lavender!14}
\cellcolor{Qwen25Lavender!50}\textbf{\tiny Q2.5} & 72 & MetaCLIP 2 ViT-M/16 (224) & 34.60 & 58.30 & 58.03 & 64.20 & 10.70 & 10.65 & 19.66 & 79.69 & 46.65 & 62.70 & 31.53 & 43.34 \\
\rowcolor{Qwen25Lavender!14}
\cellcolor{Qwen25Lavender!50}\textbf{\tiny Q2.5} & 73 & MetaCLIP 2 ViT-M/16 (mT5, 224) & 34.00 & 55.98 & 57.54 & 62.00 & 11.00 & 11.56 & 20.84 & 82.33 & 47.46 & 60.32 & 32.71 & 43.25 \\
\rowcolor{SmolPeach!16}
\cellcolor{SmolPeach!55}\textbf{\tiny Smol} & 74 & MetaCLIP ViT-G/14 (2.5B, 224) & 28.30 & 59.46 & 54.70 & 59.50 & 11.40 & 11.03 & 27.56 & 67.91 & 43.09 & \textbf{68.64} & 43.54 & 43.19 \\
\rowcolor{Qwen25Lavender!14}
\cellcolor{Qwen25Lavender!50}\textbf{\tiny Q2.5} & 75 & DINOv2 ViT-L/14 & 33.10 & 54.83 & 55.61 & 60.90 & 10.80 & 7.93 & 10.77 & 89.85 & 49.80 & 65.57 & 34.76 & 43.08 \\
\rowcolor{Qwen25Lavender!14}
\cellcolor{Qwen25Lavender!50}\textbf{\tiny Q2.5} & 76 & MetaCLIP 2 ViT-B/32 (384) & 33.90 & 57.53 & 57.42 & 62.70 & 10.70 & 11.57 & 20.97 & 76.26 & 46.34 & 61.37 & 35.02 & 43.07 \\
\rowcolor{SmolPeach!16}
\cellcolor{SmolPeach!55}\textbf{\tiny Smol} & 77 & SigLIP2 ViT-G/16 (384) & 27.40 & 56.37 & 58.99 & 57.70 & 13.60 & 18.24 & \textbf{43.18} & 64.28 & 46.14 & 54.23 & 32.51 & 42.97 \\
\rowcolor{Qwen3Mint!12}
\cellcolor{Qwen3Mint!45}\textbf{\tiny Q3} & 78 & MetaCLIP 2 ViT-M/16 (384) & 33.00 & 54.44 & 56.78 & 65.50 & 11.90 & 12.16 & 24.39 & 71.78 & 47.66 & 61.70 & 32.92 & 42.93 \\
\rowcolor{SmolPeach!16}
\cellcolor{SmolPeach!55}\textbf{\tiny Smol} & 79 & SigLIP2 So400m/14 (224) & 30.00 & 57.92 & 59.60 & 54.10 & 10.90 & 14.67 & 35.40 & 66.67 & 48.98 & 57.49 & 35.23 & 42.81 \\
\rowcolor{Qwen3Mint!12}
\cellcolor{Qwen3Mint!45}\textbf{\tiny Q3} & 80 & SigLIP2 ViT-B/16 (384) & 32.20 & 54.44 & 52.51 & 65.00 & 11.80 & 10.09 & 21.30 & 74.80 & 45.93 & 64.34 & 35.40 & 42.53 \\
\rowcolor{Qwen25Lavender!14}
\cellcolor{Qwen25Lavender!50}\textbf{\tiny Q2.5} & 81 & UniTok (256) & 33.80 & 56.76 & 56.80 & 62.30 & 11.00 & 8.44 & 12.38 & 77.89 & 48.88 & 64.44 & 34.85 & 42.50 \\
\rowcolor{Qwen3Mint!12}
\cellcolor{Qwen3Mint!45}\textbf{\tiny Q3} & 82 & SigLIP2 ViT-B/32 (256) & 33.00 & 57.14 & 57.62 & 64.20 & 10.70 & 8.91 & 18.46 & 81.65 & 47.66 & 58.04 & 30.08 & 42.50 \\
\rowcolor{SmolPeach!16}
\cellcolor{SmolPeach!55}\textbf{\tiny Smol} & 83 & PE-Lang-L/14 (448) & 26.30 & 49.42 & 56.58 & 54.00 & 13.20 & 16.90 & 36.91 & 68.13 & 47.76 & 62.23 & 35.85 & 42.48 \\
\rowcolor{SmolPeach!16}
\cellcolor{SmolPeach!55}\textbf{\tiny Smol} & 84 & PE-Core-G/14 (448) & 27.40 & 54.44 & 55.26 & 56.30 & 13.30 & 14.64 & 34.19 & 81.35 & 45.33 & 49.85 & 34.44 & 42.41 \\
\rowcolor{SmolPeach!16}
\cellcolor{SmolPeach!55}\textbf{\tiny Smol} & 85 & SigLIP2 So400m/16 (512) & 29.80 & 57.92 & 58.97 & 53.10 & 13.60 & \textbf{19.28} & 40.89 & 66.28 & 47.26 & 47.27 & 31.72 & 42.37 \\
\rowcolor{SmolPeach!16}
\cellcolor{SmolPeach!55}\textbf{\tiny Smol} & 86 & SigLIP2 ViT-L/16 (256) & 27.70 & \textbf{61.00} & 57.23 & 53.40 & 10.90 & 14.60 & 33.08 & 67.71 & 47.46 & 58.26 & 34.56 & 42.35 \\
\rowcolor{SmolPeach!16}
\cellcolor{SmolPeach!55}\textbf{\tiny Smol} & 87 & SigLIP2 ViT-B/16 (384) & 29.40 & 50.97 & 54.94 & 54.80 & 12.10 & 10.24 & 31.11 & 69.35 & 47.36 & 61.30 & 42.27 & 42.17 \\
\rowcolor{SmolPeach!16}
\cellcolor{SmolPeach!55}\textbf{\tiny Smol} & 88 & SigLIP2 So400m/16 (256) & 29.10 & 56.37 & 57.74 & 54.30 & 12.00 & 14.14 & 34.23 & 66.12 & 46.75 & 55.66 & 34.95 & 41.94 \\
\rowcolor{Qwen25Lavender!14}
\cellcolor{Qwen25Lavender!50}\textbf{\tiny Q2.5} & 89 & MetaCLIP 2 ViT-B/32 (mT5, 224) & 33.50 & 55.21 & 55.88 & 60.50 & 10.60 & 9.83 & 14.50 & 73.32 & 49.39 & 61.65 & 33.48 & 41.62 \\
\rowcolor{Qwen25Lavender!14}
\cellcolor{Qwen25Lavender!50}\textbf{\tiny Q2.5} & 90 & MetaCLIP 2 ViT-S/16 (384) & 35.20 & 56.76 & 55.39 & 60.10 & 12.00 & 12.19 & 21.98 & 72.38 & 46.34 & 55.94 & 29.08 & 41.58 \\
\rowcolor{Qwen25Lavender!14}
\cellcolor{Qwen25Lavender!50}\textbf{\tiny Q2.5} & 91 & EUPE ViT-B & 33.60 & 55.98 & 53.53 & 62.60 & 11.00 & 9.01 & 11.35 & 77.61 & 47.05 & 61.70 & 33.26 & 41.52 \\
\rowcolor{Qwen3Mint!12}
\cellcolor{Qwen3Mint!45}\textbf{\tiny Q3} & 92 & UniTok (256) & 33.50 & 50.58 & 56.01 & 65.10 & 12.50 & 9.00 & 16.84 & 80.48 & 47.15 & 53.66 & 31.42 & 41.48 \\
\rowcolor{SmolPeach!16}
\cellcolor{SmolPeach!55}\textbf{\tiny Smol} & 93 & MetaCLIP 2 ViT-L/14 (224) & 28.30 & 52.90 & 54.79 & 54.00 & 11.40 & 9.01 & 28.90 & 66.12 & 42.58 & 67.06 & 40.26 & 41.39 \\
\rowcolor{Qwen25Lavender!14}
\cellcolor{Qwen25Lavender!50}\textbf{\tiny Q2.5} & 94 & MetaCLIP 2 ViT-B/32 (224) & 33.70 & 56.37 & 55.50 & 62.10 & 11.30 & 9.05 & 16.19 & 70.12 & 47.76 & 61.08 & 31.70 & 41.35 \\
\rowcolor{Qwen25Lavender!14}
\cellcolor{Qwen25Lavender!50}\textbf{\tiny Q2.5} & 95 & MetaCLIP 2 ViT-S/16 (mT5, 224) & 33.70 & 58.69 & 55.07 & 61.10 & 10.80 & 9.48 & 16.90 & 73.82 & 46.95 & 58.74 & 29.61 & 41.35 \\
\rowcolor{Qwen25Lavender!14}
\cellcolor{Qwen25Lavender!50}\textbf{\tiny Q2.5} & 96 & DINOv2 ViT-B/14 & 35.30 & 50.97 & 55.42 & 61.60 & 11.10 & 8.12 & 10.74 & 74.70 & 46.75 & 66.29 & 33.42 & 41.31 \\
\rowcolor{Qwen25Lavender!14}
\cellcolor{Qwen25Lavender!50}\textbf{\tiny Q2.5} & 97 & EUPE ViT-S & 33.50 & 55.60 & 55.24 & 62.90 & 11.10 & 8.26 & 11.18 & 72.29 & 48.07 & 62.17 & 31.86 & 41.11 \\
\rowcolor{Qwen25Lavender!14}
\cellcolor{Qwen25Lavender!50}\textbf{\tiny Q2.5} & 98 & MetaCLIP 2 ViT-S/16 (224) & 34.70 & 56.76 & 54.27 & 59.70 & 10.90 & 10.20 & 17.68 & 76.90 & 48.78 & 54.06 & 27.61 & 41.05 \\
\rowcolor{Qwen25Lavender!14}
\cellcolor{Qwen25Lavender!50}\textbf{\tiny Q2.5} & 99 & Pixio ViT-B/16 & 32.40 & 52.90 & 53.84 & 63.10 & 10.70 & 8.66 & 11.07 & 77.48 & 46.75 & 61.05 & 33.03 & 41.00 \\
\rowcolor{Qwen3Mint!12}
\cellcolor{Qwen3Mint!45}\textbf{\tiny Q3} & 100 & MetaCLIP 2 ViT-S/16 (384) & 32.30 & 50.58 & 54.95 & 61.90 & 11.60 & 10.76 & 23.21 & 83.27 & 45.33 & 49.60 & 26.55 & 40.91 \\
\rowcolor{SmolPeach!16}
\cellcolor{SmolPeach!55}\textbf{\tiny Smol} & 101 & MetaCLIP 2 ViT-G/14 (224) & 28.80 & 55.60 & 56.74 & 55.40 & 10.40 & 8.48 & 21.87 & 72.32 & 44.00 & 58.82 & 37.42 & 40.90 \\
\rowcolor{Qwen25Lavender!14}
\cellcolor{Qwen25Lavender!50}\textbf{\tiny Q2.5} & 102 & UniAR-BSQ & 35.50 & 49.81 & 53.20 & 65.10 & 10.20 & 8.43 & 10.97 & 82.84 & 46.14 & 55.32 & 32.10 & 40.87 \\
\rowcolor{Qwen25Lavender!14}
\cellcolor{Qwen25Lavender!50}\textbf{\tiny Q2.5} & 103 & TokLIP-S (256) & 35.30 & 50.97 & 52.13 & 64.20 & 10.50 & 8.14 & 10.67 & 76.82 & 47.26 & 61.05 & 31.34 & 40.76 \\
\rowcolor{SmolPeach!16}
\cellcolor{SmolPeach!55}\textbf{\tiny Smol} & 104 & SigLIP2 ViT-B/16 (512) & 28.30 & 50.58 & 55.34 & 51.20 & 12.20 & 14.84 & 31.30 & 68.48 & 45.12 & 53.37 & 37.16 & 40.72 \\
\rowcolor{SmolPeach!16}
\cellcolor{SmolPeach!55}\textbf{\tiny Smol} & 105 & MetaCLIP ViT-L/14 (2.5B, 224) & 29.10 & 55.21 & 56.34 & 54.60 & 11.00 & 11.96 & 26.52 & 70.51 & 42.48 & 53.95 & 35.81 & 40.68 \\
\rowcolor{Qwen25Lavender!14}
\cellcolor{Qwen25Lavender!50}\textbf{\tiny Q2.5} & 106 & VILA-U (256) & 34.30 & 52.12 & 55.11 & 64.50 & 10.40 & 8.53 & 11.10 & 79.94 & 46.24 & 55.23 & 28.37 & 40.53 \\
\rowcolor{Qwen3Mint!12}
\cellcolor{Qwen3Mint!45}\textbf{\tiny Q3} & 107 & MetaCLIP 2 ViT-H/14 (378) & 31.60 & 52.12 & 50.82 & 62.50 & 10.50 & 8.44 & 13.47 & 83.99 & 43.39 & 56.97 & 31.87 & 40.52 \\
\rowcolor{SmolPeach!16}
\cellcolor{SmolPeach!55}\textbf{\tiny Smol} & 108 & OpenAI CLIP ViT-L/14 (224) & 29.00 & 57.14 & 53.83 & 56.50 & 12.00 & 10.99 & 24.99 & 66.59 & 43.80 & 56.60 & 34.19 & 40.51 \\
\rowcolor{Qwen3Mint!12}
\cellcolor{Qwen3Mint!45}\textbf{\tiny Q3} & 109 & DINOv2 ViT-G/14 & 32.50 & 52.51 & 57.88 & 61.60 & 10.50 & 7.63 & 9.97 & 87.46 & 47.87 & 50.44 & 26.91 & 40.48 \\
\rowcolor{Qwen3Mint!12}
\cellcolor{Qwen3Mint!45}\textbf{\tiny Q3} & 110 & MetaCLIP 2 ViT-L/14 (224) & 33.40 & 47.88 & 50.31 & 63.70 & 9.70 & 8.70 & 12.01 & 82.90 & 47.15 & 57.41 & 31.61 & 40.43 \\
\rowcolor{Qwen3Mint!12}
\cellcolor{Qwen3Mint!45}\textbf{\tiny Q3} & 111 & MetaCLIP 2 ViT-M/16 (mT5, 224) & 31.20 & 47.88 & 52.66 & 62.60 & 10.90 & 9.59 & 18.68 & 83.79 & 47.87 & 53.41 & 26.03 & 40.42 \\
\rowcolor{SmolPeach!16}
\cellcolor{SmolPeach!55}\textbf{\tiny Smol} & 112 & MetaCLIP ViT-H/14 (2.5B, 224) & 28.70 & 53.28 & 57.87 & 55.40 & 11.90 & 12.68 & 26.31 & 65.31 & 45.43 & 54.07 & 31.95 & 40.26 \\
\rowcolor{SmolPeach!16}
\cellcolor{SmolPeach!55}\textbf{\tiny Smol} & 113 & Web-SSL DINO 1B (224) & 28.90 & 47.88 & 55.55 & 53.00 & 12.20 & 7.71 & 10.81 & 73.91 & 48.27 & 65.84 & 35.63 & 39.97 \\
\rowcolor{Qwen3Mint!12}
\cellcolor{Qwen3Mint!45}\textbf{\tiny Q3} & 114 & PE-Lang-L/14 (448) & 30.40 & 51.35 & 52.25 & 59.20 & 10.10 & 7.01 & 12.49 & 79.24 & 46.65 & 55.89 & 34.32 & 39.90 \\
\rowcolor{Qwen3Mint!12}
\cellcolor{Qwen3Mint!45}\textbf{\tiny Q3} & 115 & Pixio ViT-L/16 & 30.50 & 50.19 & 50.31 & 64.10 & 9.80 & 7.20 & 8.75 & 85.35 & 46.54 & 56.11 & 29.12 & 39.82 \\
\rowcolor{Qwen3Mint!12}
\cellcolor{Qwen3Mint!45}\textbf{\tiny Q3} & 116 & MetaCLIP 2 ViT-S/16 (224) & 32.00 & 56.76 & 52.19 & 63.40 & 10.70 & 10.07 & 16.93 & 71.01 & 46.34 & 51.90 & 26.22 & 39.77 \\
\rowcolor{Qwen3Mint!12}
\cellcolor{Qwen3Mint!45}\textbf{\tiny Q3} & 117 & DINOv3 ViT-L/16 & 33.30 & 46.72 & 48.37 & 62.40 & 10.20 & 7.74 & 10.22 & 83.08 & 42.68 & 58.51 & 30.79 & 39.46 \\
\rowcolor{Qwen3Mint!12}
\cellcolor{Qwen3Mint!45}\textbf{\tiny Q3} & 118 & VILA-U (256) & 31.50 & 48.26 & 52.16 & 65.60 & 11.10 & 7.69 & 11.45 & 83.84 & 45.02 & 52.16 & 25.14 & 39.45 \\
\rowcolor{SmolPeach!16}
\cellcolor{SmolPeach!55}\textbf{\tiny Smol} & 119 & MetaCLIP ViT-H/14 (v1.2, 224) & 30.10 & 54.44 & 57.05 & 55.00 & 11.50 & 11.33 & 27.63 & 73.03 & 44.82 & 42.93 & 25.83 & 39.42 \\
\rowcolor{Qwen3Mint!12}
\cellcolor{Qwen3Mint!45}\textbf{\tiny Q3} & 120 & Pixio ViT-B/16 & 31.30 & 47.49 & 50.23 & 64.70 & 10.10 & 8.52 & 9.69 & 80.94 & 44.11 & 56.81 & 29.67 & 39.41 \\
\rowcolor{SmolPeach!16}
\cellcolor{SmolPeach!55}\textbf{\tiny Smol} & 121 & SigLIP2 ViT-B/16 (256) & 28.50 & 51.35 & 55.81 & 52.90 & 9.80 & 11.05 & 27.68 & 66.43 & 48.27 & 51.34 & 28.06 & 39.20 \\
\rowcolor{Qwen3Mint!12}
\cellcolor{Qwen3Mint!45}\textbf{\tiny Q3} & 122 & SigLIP2 ViT-L/16 (512) & 31.90 & 50.58 & 50.16 & 62.10 & 10.30 & 7.88 & 12.06 & 79.03 & 43.29 & 52.14 & 31.60 & 39.19 \\
\rowcolor{Qwen3Mint!12}
\cellcolor{Qwen3Mint!45}\textbf{\tiny Q3} & 123 & MetaCLIP 2 ViT-S/16 (mT5, 224) & 32.60 & 49.03 & 51.02 & 63.90 & 11.00 & 9.42 & 15.76 & 73.85 & 44.21 & 52.35 & 27.17 & 39.12 \\
\rowcolor{Qwen25Lavender!14}
\cellcolor{Qwen25Lavender!50}\textbf{\tiny Q2.5} & 124 & MetaCLIP ViT-B/32 (2.5B, 224) & 34.70 & 50.58 & 49.24 & 62.90 & 11.00 & 7.82 & 10.79 & 75.34 & 45.93 & 54.87 & 26.24 & 39.04 \\
\rowcolor{Qwen3Mint!12}
\cellcolor{Qwen3Mint!45}\textbf{\tiny Q3} & 125 & TokLIP-S (256) & 31.40 & 53.28 & 52.43 & 64.90 & 10.90 & 7.55 & 11.53 & 85.97 & 45.93 & 41.63 & 23.75 & 39.02 \\
\rowcolor{SmolPeach!16}
\cellcolor{SmolPeach!55}\textbf{\tiny Smol} & 126 & SigLIP2 ViT-B/16 (224) & 27.10 & 55.60 & 56.02 & 59.40 & 10.10 & 10.65 & 23.12 & 66.28 & 45.22 & 41.34 & 33.09 & 38.90 \\
\rowcolor{SmolPeach!16}
\cellcolor{SmolPeach!55}\textbf{\tiny Smol} & 127 & MetaCLIP 2 ViT-B/16 (384) & 30.00 & 48.26 & 53.60 & 50.00 & 10.80 & 12.13 & 26.22 & 68.46 & 44.72 & 54.17 & 28.47 & 38.80 \\
\rowcolor{Qwen25Lavender!14}
\cellcolor{Qwen25Lavender!50}\textbf{\tiny Q2.5} & 128 & DINOv2 ViT-S/14 & 32.20 & 45.17 & 50.86 & 61.90 & 10.70 & 8.41 & 9.94 & 80.90 & 46.75 & 54.46 & 25.29 & 38.78 \\
\rowcolor{Qwen3Mint!12}
\cellcolor{Qwen3Mint!45}\textbf{\tiny Q3} & 129 & MetaCLIP 2 ViT-M/16 (224) & 32.10 & 49.81 & 53.08 & 65.00 & 11.20 & 8.13 & 15.68 & 78.56 & 45.73 & 43.42 & 23.70 & 38.76 \\
\rowcolor{Qwen3Mint!12}
\cellcolor{Qwen3Mint!45}\textbf{\tiny Q3} & 130 & MetaCLIP ViT-B/32 (2.5B, 224) & 31.60 & 49.81 & 47.85 & 65.40 & 10.60 & 8.30 & 13.91 & 78.20 & 41.87 & 50.28 & 24.93 & 38.43 \\
\rowcolor{Qwen3Mint!12}
\cellcolor{Qwen3Mint!45}\textbf{\tiny Q3} & 131 & MetaCLIP 2 ViT-B/32 (224) & 34.40 & 46.72 & 49.04 & 64.00 & 11.00 & 8.33 & 13.17 & 83.28 & 43.39 & 46.02 & 21.34 & 38.24 \\
\rowcolor{Qwen3Mint!12}
\cellcolor{Qwen3Mint!45}\textbf{\tiny Q3} & 132 & TokLIP-L (384) & 33.10 & 46.72 & 48.94 & 61.70 & 11.50 & 7.93 & 11.68 & 80.29 & 44.41 & 50.09 & 24.14 & 38.23 \\
\rowcolor{Qwen3Mint!12}
\cellcolor{Qwen3Mint!45}\textbf{\tiny Q3} & 133 & EUPE ConvNeXt-B & 32.10 & 43.63 & 47.17 & 62.60 & 10.20 & 7.97 & 10.29 & 77.71 & 42.68 & 51.41 & 29.00 & 37.71 \\
\rowcolor{Qwen25Lavender!14}
\cellcolor{Qwen25Lavender!50}\textbf{\tiny Q2.5} & 134 & MetaCLIP ViT-B/32 (400M, 224) & 34.70 & 49.03 & 47.62 & 62.70 & 10.20 & 7.79 & 10.95 & 71.55 & 43.29 & 49.98 & 26.10 & 37.63 \\
\rowcolor{SmolPeach!16}
\cellcolor{SmolPeach!55}\textbf{\tiny Smol} & 135 & MetaCLIP 2 ViT-M/16 (384) & 28.80 & 42.47 & 52.38 & 50.80 & 11.40 & 10.18 & 21.49 & 69.18 & 41.16 & 54.44 & 31.33 & 37.60 \\
\rowcolor{SmolPeach!16}
\cellcolor{SmolPeach!55}\textbf{\tiny Smol} & 136 & MetaCLIP ViT-L/14 (400M, 224) & 29.80 & 51.35 & 52.77 & 54.20 & 10.90 & 10.03 & 23.44 & 64.31 & 41.67 & 46.45 & 28.30 & 37.57 \\
\rowcolor{Qwen3Mint!12}
\cellcolor{Qwen3Mint!45}\textbf{\tiny Q3} & 137 & EUPE ViT-S & 32.80 & 43.24 & 46.21 & 62.80 & 10.60 & 7.43 & 9.39 & 70.44 & 44.21 & 56.39 & 29.14 & 37.51 \\
\rowcolor{SmolPeach!16}
\cellcolor{SmolPeach!55}\textbf{\tiny Smol} & 138 & DINOv2 ViT-L/14 & \textbf{31.20} & 44.02 & 49.54 & 55.10 & 10.80 & 7.93 & 9.84 & 77.97 & 42.78 & 54.61 & 28.46 & 37.48 \\
\rowcolor{SmolPeach!16}
\cellcolor{SmolPeach!55}\textbf{\tiny Smol} & 139 & SigLIP2 ViT-B/32 (256) & 27.10 & 55.98 & 53.83 & 53.00 & 10.80 & 8.81 & 17.11 & 69.10 & 44.41 & 45.01 & 24.24 & 37.22 \\
\rowcolor{SmolPeach!16}
\cellcolor{SmolPeach!55}\textbf{\tiny Smol} & 140 & MetaCLIP ViT-B/16 (2.5B, 224) & 28.40 & 49.81 & 49.77 & 54.40 & 10.60 & 10.08 & 19.84 & 66.51 & 41.16 & 50.75 & 26.25 & 37.05 \\
\rowcolor{SmolPeach!16}
\cellcolor{SmolPeach!55}\textbf{\tiny Smol} & 141 & TokLIP-L (384) & 27.30 & 52.51 & 51.66 & 56.90 & 11.20 & 6.88 & 13.05 & 82.62 & 44.41 & 35.25 & 25.02 & 36.98 \\
\rowcolor{SmolPeach!16}
\cellcolor{SmolPeach!55}\textbf{\tiny Smol} & 142 & Pixio ViT-H/16 & 29.10 & 42.86 & 48.46 & 53.30 & 10.10 & 7.28 & 10.16 & 78.16 & 43.80 & 55.76 & 27.54 & 36.96 \\
\rowcolor{SmolPeach!16}
\cellcolor{SmolPeach!55}\textbf{\tiny Smol} & 143 & MetaCLIP ViT-B/16 (400M, 224) & 27.90 & 47.10 & 51.42 & 55.00 & 11.50 & 8.39 & 19.16 & 71.43 & 40.65 & 47.57 & 25.72 & 36.89 \\
\rowcolor{Qwen25Lavender!14}
\cellcolor{Qwen25Lavender!50}\textbf{\tiny Q2.5} & 144 & DINOv3 ViT-L/16 & 32.20 & 44.02 & 50.84 & 61.90 & 10.70 & 8.61 & 10.88 & 71.63 & 43.19 & 48.48 & 23.26 & 36.88 \\
\rowcolor{SmolPeach!16}
\cellcolor{SmolPeach!55}\textbf{\tiny Smol} & 145 & MetaCLIP 2 ViT-B/16 (224) & 28.00 & 48.26 & 51.36 & 55.10 & 11.60 & 9.60 & 21.12 & 68.38 & 39.74 & 46.99 & 25.58 & 36.88 \\
\rowcolor{SmolPeach!16}
\cellcolor{SmolPeach!55}\textbf{\tiny Smol} & 146 & EUPE ConvNeXt-B & 30.60 & 50.97 & 52.37 & 53.10 & 9.60 & 5.76 & 10.88 & 67.22 & 41.97 & 52.32 & 29.24 & 36.73 \\
\rowcolor{SmolPeach!16}
\cellcolor{SmolPeach!55}\textbf{\tiny Smol} & 147 & DINOv3 ViT-L/16 & 29.90 & 44.79 & 49.10 & 53.10 & 10.90 & 8.07 & 11.88 & 69.69 & 42.89 & 50.91 & 31.25 & 36.59 \\
\rowcolor{Qwen3Mint!12}
\cellcolor{Qwen3Mint!45}\textbf{\tiny Q3} & 148 & I-JEPA ViT-H/14 & 30.80 & 41.70 & 47.08 & 63.20 & 10.80 & 8.01 & 8.91 & 82.99 & 42.68 & 46.11 & 19.83 & 36.56 \\
\rowcolor{Qwen25Lavender!14}
\cellcolor{Qwen25Lavender!50}\textbf{\tiny Q2.5} & 149 & I-JEPA ViT-H/14 & 31.70 & 42.08 & 47.97 & 62.00 & 10.10 & 7.64 & 9.91 & 72.29 & 44.31 & 52.83 & 20.45 & 36.48 \\
\rowcolor{Qwen3Mint!12}
\cellcolor{Qwen3Mint!45}\textbf{\tiny Q3} & 150 & DINOv2 ViT-S/14 & 31.10 & 41.31 & 47.18 & 61.50 & 10.80 & 7.75 & 9.32 & 75.04 & 44.41 & 48.22 & 23.06 & 36.34 \\
\rowcolor{Qwen3Mint!12}
\cellcolor{Qwen3Mint!45}\textbf{\tiny Q3} & 151 & DINOv2 ViT-L/14 & 30.50 & 38.61 & 47.02 & 62.50 & 10.20 & 8.56 & 9.11 & 74.44 & 42.89 & 50.13 & 25.05 & 36.27 \\
\rowcolor{Qwen25Lavender!14}
\cellcolor{Qwen25Lavender!50}\textbf{\tiny Q2.5} & 152 & RAEv2 DINOv3-L & 33.80 & 48.26 & 48.33 & 61.60 & 11.10 & 8.36 & 11.47 & 66.59 & 40.35 & 47.03 & 22.01 & 36.26 \\
\rowcolor{Qwen25Lavender!14}
\cellcolor{Qwen25Lavender!50}\textbf{\tiny Q2.5} & 153 & Web-SSL MAE 3B (224) & 33.50 & 44.02 & 49.55 & 62.90 & 10.80 & 7.54 & 9.39 & 70.47 & 44.00 & 45.03 & 20.27 & 36.13 \\
\rowcolor{Qwen25Lavender!14}
\cellcolor{Qwen25Lavender!50}\textbf{\tiny Q2.5} & 154 & Pixio ViT-H/16 & 32.80 & 25.10 & 50.07 & 60.70 & 11.10 & 8.27 & 10.14 & 65.49 & 43.19 & 58.44 & 31.44 & 36.07 \\
\rowcolor{SmolPeach!16}
\cellcolor{SmolPeach!55}\textbf{\tiny Smol} & 155 & SigLIP2 ViT-L/16 (512) & 27.70 & 46.33 & 49.06 & 54.60 & 10.80 & 7.88 & 11.57 & 68.33 & 41.57 & 49.99 & 26.23 & 35.82 \\
\rowcolor{SmolPeach!16}
\cellcolor{SmolPeach!55}\textbf{\tiny Smol} & 156 & PE-Core-B/16 (224) & 26.10 & 42.86 & 50.35 & 49.70 & 10.40 & 8.80 & 19.58 & 73.85 & 45.22 & 41.92 & 24.70 & 35.77 \\
\rowcolor{SmolPeach!16}
\cellcolor{SmolPeach!55}\textbf{\tiny Smol} & 157 & EUPE ViT-B & 27.90 & 51.74 & 49.11 & 50.40 & 9.60 & 8.48 & 12.85 & 64.61 & 41.67 & 48.59 & 28.20 & 35.74 \\
\rowcolor{SmolPeach!16}
\cellcolor{SmolPeach!55}\textbf{\tiny Smol} & 158 & MetaCLIP 2 ViT-B/32 (384) & 29.20 & 42.08 & 51.49 & 50.70 & 10.70 & 9.05 & 17.94 & 62.57 & 43.50 & 47.61 & 24.60 & 35.40 \\
\rowcolor{Qwen3Mint!12}
\cellcolor{Qwen3Mint!45}\textbf{\tiny Q3} & 159 & MetaCLIP 2 ViT-B/32 (mT5, 224) & 31.70 & 37.84 & 45.29 & 64.00 & 10.50 & 8.26 & 9.60 & 72.11 & 38.41 & 47.49 & 24.06 & 35.39 \\
\rowcolor{SmolPeach!16}
\cellcolor{SmolPeach!55}\textbf{\tiny Smol} & 160 & DINOv2 ViT-G/14 & 29.10 & 44.02 & 50.30 & 52.00 & 10.60 & 5.75 & 10.07 & 67.41 & 41.77 & 46.95 & 26.96 & 34.99 \\
\rowcolor{Qwen3Mint!12}
\cellcolor{Qwen3Mint!45}\textbf{\tiny Q3} & 161 & UniAR-BSQ & 31.40 & 44.79 & 48.24 & 63.50 & 10.90 & 7.71 & 9.46 & 69.45 & 41.46 & 36.32 & 20.62 & 34.90 \\
\rowcolor{SmolPeach!16}
\cellcolor{SmolPeach!55}\textbf{\tiny Smol} & 162 & MetaCLIP 2 ViT-H/14 (378) & 30.30 & 53.28 & 49.55 & 53.20 & 12.20 & 8.33 & 19.93 & 46.47 & 42.89 & 44.08 & 23.43 & 34.88 \\
\rowcolor{Qwen25Lavender!14}
\cellcolor{Qwen25Lavender!50}\textbf{\tiny Q2.5} & 163 & DINO ViT-B/16 & 33.30 & 39.38 & 45.10 & 63.00 & 10.30 & 8.14 & 9.39 & 75.14 & 43.50 & 40.55 & 15.49 & 34.84 \\
\rowcolor{Qwen3Mint!12}
\cellcolor{Qwen3Mint!45}\textbf{\tiny Q3} & 164 & EUPE ViT-T & 33.70 & 40.15 & 44.48 & 63.00 & 9.30 & 7.81 & 8.74 & 81.99 & 39.23 & 36.11 & 17.63 & 34.74 \\
\rowcolor{Qwen3Mint!12}
\cellcolor{Qwen3Mint!45}\textbf{\tiny Q3} & 165 & MetaCLIP ViT-B/32 (400M, 224) & 33.60 & 39.38 & 45.84 & 63.40 & 10.60 & 6.83 & 10.46 & 74.31 & 39.53 & 38.13 & 19.49 & 34.69 \\
\rowcolor{Qwen3Mint!12}
\cellcolor{Qwen3Mint!45}\textbf{\tiny Q3} & 166 & EUPE ViT-B & 31.10 & 37.84 & 45.59 & 61.40 & 10.40 & 7.61 & 10.92 & 68.73 & 42.28 & 42.92 & 22.02 & 34.62 \\
\rowcolor{SmolPeach!16}
\cellcolor{SmolPeach!55}\textbf{\tiny Smol} & 167 & MetaCLIP 2 ViT-B/32 (mT5, 224) & 28.70 & 46.33 & 49.50 & 54.30 & 10.20 & 6.91 & 13.78 & 68.43 & 44.82 & 37.35 & 20.13 & 34.59 \\
\rowcolor{SmolPeach!16}
\cellcolor{SmolPeach!55}\textbf{\tiny Smol} & 168 & Pixio ViT-L/16 & 26.90 & 40.15 & 49.98 & 54.90 & 9.50 & 7.30 & 9.89 & 69.37 & 40.75 & 45.71 & 25.86 & 34.57 \\
\rowcolor{SmolPeach!16}
\cellcolor{SmolPeach!55}\textbf{\tiny Smol} & 169 & MetaCLIP 2 ViT-M/16 (mT5, 224) & 27.40 & 44.40 & 46.92 & 49.00 & 10.80 & 8.52 & 18.31 & 65.96 & 37.09 & 47.85 & 23.71 & 34.54 \\
\rowcolor{SmolPeach!16}
\cellcolor{SmolPeach!55}\textbf{\tiny Smol} & 170 & MetaCLIP 2 ViT-M/16 (224) & 29.30 & 46.72 & 49.79 & 54.60 & 10.60 & 8.57 & 17.13 & 71.09 & 39.94 & 33.94 & 17.51 & 34.47 \\
\rowcolor{SmolPeach!16}
\cellcolor{SmolPeach!55}\textbf{\tiny Smol} & 171 & TokLIP-S (256) & 28.30 & 48.26 & 49.28 & 55.00 & 11.30 & 6.48 & 10.46 & 66.28 & 39.63 & 41.66 & 22.49 & 34.47 \\
\rowcolor{SmolPeach!16}
\cellcolor{SmolPeach!55}\textbf{\tiny Smol} & 172 & MetaCLIP ViT-B/32 (2.5B, 224) & 28.80 & 43.24 & 46.19 & 52.60 & 9.80 & 8.29 & 12.54 & 68.51 & 38.62 & 44.31 & 21.75 & 34.06 \\
\rowcolor{SmolPeach!16}
\cellcolor{SmolPeach!55}\textbf{\tiny Smol} & 173 & MetaCLIP 2 ViT-B/32 (224) & 25.50 & 44.40 & 47.58 & 53.90 & 10.40 & 7.43 & 13.23 & 66.75 & 40.45 & 45.29 & 19.65 & 34.05 \\
\rowcolor{Qwen3Mint!12}
\cellcolor{Qwen3Mint!45}\textbf{\tiny Q3} & 174 & DINOv2 ViT-B/14 & 30.50 & 36.68 & 44.81 & 61.20 & 10.60 & 7.06 & 9.37 & 68.12 & 40.85 & 44.31 & 20.30 & 33.98 \\
\rowcolor{Qwen25Lavender!14}
\cellcolor{Qwen25Lavender!50}\textbf{\tiny Q2.5} & 175 & Web-SSL MAE 1B (224) & 33.10 & 37.84 & 45.12 & 61.20 & 10.60 & 8.00 & 10.12 & 70.41 & 41.97 & 37.20 & 15.44 & 33.73 \\
\rowcolor{Qwen3Mint!12}
\cellcolor{Qwen3Mint!45}\textbf{\tiny Q3} & 176 & Pixio ViT-H/16 & 29.90 & 29.34 & 43.98 & 61.10 & 9.80 & 7.56 & 10.00 & 68.43 & 40.96 & 45.55 & 23.70 & 33.67 \\
\rowcolor{Qwen25Lavender!14}
\cellcolor{Qwen25Lavender!50}\textbf{\tiny Q2.5} & 177 & Web-SSL MAE 300M (224) & 33.60 & 36.68 & 45.35 & 62.30 & 10.60 & 7.71 & 7.94 & 71.34 & 41.06 & 36.69 & 16.23 & 33.59 \\
\rowcolor{Qwen3Mint!12}
\cellcolor{Qwen3Mint!45}\textbf{\tiny Q3} & 178 & RAEv2 DINOv3-L& 30.40 & 40.15 & 44.83 & 60.20 & 10.00 & 8.50 & 10.64 & 67.14 & 39.74 & 38.55 & 18.58 & 33.52 \\
\rowcolor{SmolPeach!16}
\cellcolor{SmolPeach!55}\textbf{\tiny Smol} & 179 & MetaCLIP 2 ViT-S/16 (mT5, 224) & 26.00 & 38.61 & 48.29 & 49.90 & 9.80 & 9.01 & 14.60 & 75.74 & 42.17 & 36.06 & 18.05 & 33.48 \\
\rowcolor{Qwen3Mint!12}
\cellcolor{Qwen3Mint!45}\textbf{\tiny Q3} & 180 & DINO ViT-B/16 & 34.10 & 38.22 & 43.74 & 62.00 & 10.90 & 7.65 & 8.52 & 71.95 & 40.55 & 36.23 & 13.79 & 33.42 \\
\rowcolor{SmolPeach!16}
\cellcolor{SmolPeach!55}\textbf{\tiny Smol} & 181 & VILA-U (256) & 29.00 & 45.17 & 47.72 & 54.50 & 10.30 & 7.03 & 11.03 & 79.14 & 40.75 & 24.68 & 17.35 & 33.33 \\
\rowcolor{SmolPeach!16}
\cellcolor{SmolPeach!55}\textbf{\tiny Smol} & 182 & I-JEPA ViT-H/14 & 27.90 & 35.91 & 47.05 & 53.20 & 10.70 & 7.43 & 9.27 & 66.67 & 39.43 & 47.83 & 20.11 & 33.23 \\
\rowcolor{SmolPeach!16}
\cellcolor{SmolPeach!55}\textbf{\tiny Smol} & 183 & MetaCLIP 2 ViT-S/16 (384) & 28.50 & 38.22 & 48.18 & 51.50 & 11.20 & 6.89 & 17.47 & 65.65 & 39.02 & 34.54 & 23.58 & 33.16 \\
\rowcolor{SmolPeach!16}
\cellcolor{SmolPeach!55}\textbf{\tiny Smol} & 184 & MetaCLIP ViT-B/32 (400M, 224) & 29.40 & 37.84 & 45.76 & 54.60 & 9.60 & 7.98 & 13.51 & 66.36 & 38.92 & 39.68 & 20.62 & 33.12 \\
\rowcolor{SmolPeach!16}
\cellcolor{SmolPeach!55}\textbf{\tiny Smol} & 185 & Pixio ViT-B/16 & 29.30 & 43.24 & 49.84 & 53.30 & 10.10 & 7.47 & 9.13 & 66.36 & 40.55 & 36.60 & 18.28 & 33.11 \\
\rowcolor{SmolPeach!16}
\cellcolor{SmolPeach!55}\textbf{\tiny Smol} & 186 & RAEv2 DINOv3-L & 26.20 & 38.22 & 46.29 & 47.10 & 7.20 & 5.09 & 7.37 & 77.07 & 39.23 & 46.01 & 24.25 & 33.09 \\
\rowcolor{SmolPeach!16}
\cellcolor{SmolPeach!55}\textbf{\tiny Smol} & 187 & EUPE ViT-S & 28.40 & 41.31 & 45.10 & 49.70 & 10.10 & 5.67 & 9.18 & 67.63 & 37.50 & 43.60 & 22.52 & 32.79 \\
\rowcolor{Qwen3Mint!12}
\cellcolor{Qwen3Mint!45}\textbf{\tiny Q3} & 188 & DINO ViT-S/8 & 30.70 & 33.98 & 42.55 & 60.20 & 10.60 & 6.57 & 9.50 & 74.68 & 39.43 & 35.80 & 15.01 & 32.64 \\
\rowcolor{SmolPeach!16}
\cellcolor{SmolPeach!55}\textbf{\tiny Smol} & 189 & DINOv2 ViT-S/14 & 27.70 & 37.84 & 47.37 & 51.80 & 10.50 & 6.89 & 9.09 & 69.75 & 41.16 & 35.46 & 17.24 & 32.25 \\
\rowcolor{SmolPeach!16}
\cellcolor{SmolPeach!55}\textbf{\tiny Smol} & 190 & UniTok (256) & 28.90 & 33.98 & 46.84 & 51.60 & 9.40 & 8.06 & 10.13 & 69.38 & 40.45 & 31.60 & 24.36 & 32.25 \\
\rowcolor{SmolPeach!16}
\cellcolor{SmolPeach!55}\textbf{\tiny Smol} & 191 & DINOv2 ViT-B/14 & 31.10 & 39.38 & 47.29 & 47.80 & 9.80 & 6.52 & 9.09 & 66.51 & 42.68 & 33.88 & 19.07 & 32.10 \\
\rowcolor{Qwen25Lavender!14}
\cellcolor{Qwen25Lavender!50}\textbf{\tiny Q2.5} & 192 & DINO ViT-B/8 & 32.50 & 32.82 & 43.22 & 61.90 & 10.40 & 8.22 & 8.28 & 69.27 & 40.14 & 26.83 & 11.10 & 31.33 \\
\rowcolor{SmolPeach!16}
\cellcolor{SmolPeach!55}\textbf{\tiny Smol} & 193 & DINO ViT-B/16 & 28.70 & 30.12 & 44.52 & 54.40 & 10.20 & 6.77 & 7.54 & 66.04 & 38.52 & 36.91 & 16.55 & 30.93 \\
\rowcolor{Qwen25Lavender!14}
\cellcolor{Qwen25Lavender!50}\textbf{\tiny Q2.5} & 194 & DINO ViT-S/8 & 32.30 & 25.87 & 44.07 & 62.80 & 11.10 & 8.22 & 9.87 & 66.82 & 37.40 & 26.00 & 12.99 & 30.68 \\
\rowcolor{Qwen3Mint!12}
\cellcolor{Qwen3Mint!45}\textbf{\tiny Q3} & 195 & Web-SSL MAE 3B (224) & 32.00 & 34.36 & 41.50 & 60.60 & 10.60 & 6.46 & 9.24 & 67.55 & 37.70 & 24.35 & 12.56 & 30.63 \\
\rowcolor{SmolPeach!16}
\cellcolor{SmolPeach!55}\textbf{\tiny Smol} & 196 & MetaCLIP 2 ViT-S/16 (224) & 26.10 & 36.29 & 46.78 & 49.40 & 10.40 & 8.60 & 14.84 & 66.04 & 39.53 & 32.07 & 6.48 & 30.59 \\
\rowcolor{SmolPeach!16}
\cellcolor{SmolPeach!55}\textbf{\tiny Smol} & 197 & UniAR-BSQ & 28.20 & 35.14 & 45.68 & 53.30 & 10.30 & 7.04 & 8.77 & 63.67 & 37.60 & 25.60 & 17.31 & 30.24 \\
\rowcolor{Qwen3Mint!12}
\cellcolor{Qwen3Mint!45}\textbf{\tiny Q3} & 198 & DINO ViT-S/16 & 30.20 & 27.03 & 40.97 & 62.20 & 10.30 & 7.09 & 8.48 & 68.43 & 38.21 & 24.38 & 11.93 & 29.93 \\
\rowcolor{Qwen3Mint!12}
\cellcolor{Qwen3Mint!45}\textbf{\tiny Q3} & 199 & Web-SSL MAE 1B (224) & 34.30 & 27.03 & 40.44 & 62.40 & 9.80 & 7.32 & 8.30 & 66.28 & 37.80 & 21.51 & 11.57 & 29.70 \\
\rowcolor{Qwen25Lavender!14}
\cellcolor{Qwen25Lavender!50}\textbf{\tiny Q2.5} & 200 & DINO ViT-S/16 & 32.70 & 29.73 & 41.19 & 62.40 & 11.00 & 8.54 & 9.53 & 66.36 & 38.31 & 18.81 & 8.07 & 29.69 \\
\rowcolor{SmolPeach!16}
\cellcolor{SmolPeach!55}\textbf{\tiny Smol} & 201 & EUPE ViT-T & 28.50 & 28.19 & 43.41 & 47.00 & 10.30 & 5.77 & 8.93 & 66.36 & 37.20 & 34.16 & 16.61 & 29.68 \\
\rowcolor{SmolPeach!16}
\cellcolor{SmolPeach!55}\textbf{\tiny Smol} & 202 & Web-SSL MAE 300M (224) & 27.00 & 32.05 & 42.20 & 53.70 & 10.40 & 5.82 & 7.53 & 66.28 & 36.79 & 27.74 & 13.24 & 29.34 \\
\rowcolor{Qwen3Mint!12}
\cellcolor{Qwen3Mint!45}\textbf{\tiny Q3} & 203 & DINO ViT-B/8 & 30.00 & 23.17 & 39.73 & 62.50 & 10.70 & 6.91 & 8.85 & 66.28 & 37.20 & 26.02 & 11.11 & 29.32 \\
\rowcolor{Qwen3Mint!12}
\cellcolor{Qwen3Mint!45}\textbf{\tiny Q3} & 204 & Web-SSL MAE 300M (224) & 29.60 & 27.03 & 41.79 & 61.20 & 10.10 & 7.42 & 9.46 & 68.98 & 37.40 & 14.23 & 10.17 & 28.85 \\
\rowcolor{SmolPeach!16}
\cellcolor{SmolPeach!55}\textbf{\tiny Smol} & 205 & DINO ViT-S/16 & 26.90 & 28.19 & 39.71 & 54.80 & 9.80 & 6.32 & 5.84 & 67.63 & 37.60 & 26.56 & 12.38 & 28.70 \\
\rowcolor{SmolPeach!16}
\cellcolor{SmolPeach!55}\textbf{\tiny Smol} & 206 & DINO ViT-B/8 & 27.30 & 32.05 & 38.16 & 53.40 & 10.00 & 6.51 & 7.56 & 66.82 & 35.26 & 26.77 & 10.68 & 28.59 \\
\rowcolor{Qwen25Lavender!14}
\cellcolor{Qwen25Lavender!50}\textbf{\tiny Q2.5} & 207 & EUPE ViT-T & 31.20 & 25.10 & 40.38 & 60.70 & 11.00 & 7.27 & 9.15 & 66.28 & 34.76 & 5.93 & 4.61 & 26.94 \\
\rowcolor{SmolPeach!16}
\cellcolor{SmolPeach!55}\textbf{\tiny Smol} & 208 & DINO ViT-S/8 & 26.20 & 24.71 & 36.81 & 54.30 & 10.00 & 7.39 & 6.95 & 66.91 & 35.37 & 11.09 & 6.77 & 26.05 \\
\rowcolor{SmolPeach!16}
\cellcolor{SmolPeach!55}\textbf{\tiny Smol} & 209 & Web-SSL MAE 3B (224) & 26.60 & 25.48 & 39.10 & 53.40 & 10.40 & 7.54 & 8.58 & 62.32 & 35.57 & 3.98 & 5.75 & 25.34 \\
\rowcolor{SmolPeach!16}
\cellcolor{SmolPeach!55}\textbf{\tiny Smol} & 210 & Web-SSL MAE 1B (224) & 26.60 & 23.55 & 34.19 & 50.10 & 10.50 & 7.61 & 8.16 & 66.28 & 30.39 & 1.54 & 3.28 & 23.84 \\

\end{xltabular}
\endgroup

\section{Additional Training and Evaluation Details}
\label{app:training_evaluation_details}

\noindent\textbf{Training recipe.}
All MLLMs follow a controlled two-stage LLaVA-style recipe implemented using LLaVA-NeXT~\citep{liu2024llavanext} for continuous tokenizers and an equivalent discrete adapter for discrete tokenizers. The visual tokenizer remains frozen throughout. In stage~1, referred to as \emph{mid-training} in this paper, only the two-layer MLP projector is trained on LLaVA-LCS-558K. In stage~2 (instruction tuning), the projector and language model are jointly trained on the LLaVA-1.5 665K mixture. Images are resized to square crops at each tokenizer's native resolution, using flat patch-token merging without AnyRes tiling. 
For language-supervised vision encoders (CLIP, MetaCLIP, and SigLIP2), we use the outputs of the penultimate Transformer block; for self-supervised and reconstruction-based encoders, we use the outputs of the final block.
We train each model on eight NVIDIA A100-SXM4-80GB GPUs using distributed data parallelism (DDP)~\citep{li2020pytorch} with bfloat16 precision~\citep{kalamkar2019bfloat16}. We use AdamW~\citep{loshchilov2019adamw} with cosine learning-rate decay~\citep{loshchilov2017sgdr}, zero weight decay, and gradient clipping~\citep{pascanu2013difficulty} at~$1.0$. The context length is 8{,}192, and each stage is trained for one epoch.

\noindent\textbf{Optimization.}
The effective per-GPU batch size is 128 during mid-training and 32 during instruction tuning, corresponding to global batch sizes of 1{,}024 and 256, respectively. For high-resolution tokenizers or tokenizers with large backbones, we reduce the micro-batch size and increase gradient accumulation to preserve the effective batch size. Mid-training uses a learning rate of $1{\times}10^{-3}$ and a warmup ratio of $0.05$. During instruction tuning, the learning rates are $1{\times}10^{-5}$ for the language model and $1{\times}10^{-4}$ for the projector, with a warmup ratio of $0.03$. Gradient checkpointing~\citep{chen2016training} is enabled during instruction tuning.

\noindent\textbf{Evaluation protocol.}
We evaluate each instruction-tuned checkpoint using
VLMEvalKit~\citep{duan2024vlmevalkit} on eleven benchmarks and splits:
MMMU test~\citep{yue2024mmmu}, MMBench-EN v1.1
test~\citep{liu2024mmbench}, ScienceQA val~\citep{lu2022scienceqa},
VQAv2 val~\citep{goyal2017vqav2}, GQA Test-Dev
balanced~\citep{hudson2019gqa}, ChartQA test~\citep{masry2022chartqa},
DocVQA val~\citep{mathew2021docvqa}, TextVQA
val~\citep{singh2019textvqa}, POPE~\citep{li2023pope}, and the Karpathy
test splits of MSCOCO~\citep{lin2014coco,karpathy2015deep} and
Flickr30K~\citep{young2014flickr30k,karpathy2015deep}.
For each benchmark, we sample up to 1{,}000 examples using a fixed seed of
42 and use the same subset for every MLLM. Decoding is greedy with a maximum
of 128 newly generated tokens.

\noindent\textbf{Scoring.}
Responses on MMMU and MMBench are parsed from free-form generations using a locally deployed Qwen3-32B judge. All other benchmarks use standard rule-based metrics: multiple-choice exact match for ScienceQA; official VQA soft accuracy~\citep{antol2015vqa} for VQAv2 and TextVQA; exact match for GQA; relaxed accuracy with a 5\% relative tolerance for ChartQA; ANLS~\citep{biten2019scene} for DocVQA; yes/no F1 for POPE; and CIDEr~\citep{vedantam2015cider} for the two captioning datasets. Accuracies expressed as percentages and CIDEr multiplied by 100. The aggregate score is the unweighted mean over the eleven benchmarks. Invalid or missing benchmark scores are excluded from the mean rather than imputed.

\section{Implementation Details}
\label{app:implementation_details}

\noindent\textbf{MLLM construction.}
For each vision-encoder--language-backbone pair, we connect the vision encoder
to the LLM through a separately trained two-layer MLP projector with GELU
activation~\citep{hendrycks2016gelu}.

\noindent\textbf{Calibration data and feature extraction.}
For methods requiring paired data, such as
MutualNN~\citep{huh2024platonic}, we follow \citet{li2026rethinking} and
randomly sample 1{,}000 image--text pairs from
LLaVA-LCS-558K~\citep{liu2024llava15}. We extract visual patch tokens from the
encoder's final layer and textual hidden states from the LLM's penultimate
layer.

\noindent\textbf{Default settings.}
Unless otherwise specified, we set the regularization parameter to
\(\epsilon=1\times10^{-4}\) and the neighborhood size to \(k=100\).

\noindent\textbf{Compute.}
All MLLM training and downstream evaluation are conducted on eight NVIDIA
A100 GPUs with 80\,GB of memory each.

\section{Details of evaluation metrics}
\label{app:detailed_metric}
\subsection{Prediction Quality}
\label{sec:prediction_quality}

\noindent\textbf{Spearman correlation.}
Let $s_i$ denote the score assigned by a prediction method to the $i$-th
vision encoder, and let $y_{i,t}$ denote its downstream MLLM performance on
benchmark $t$. We use Spearman's rank correlation to measure ranking quality:
\[
\rho_t =
\frac{
    \sum_{i=1}^{N}(R_i-\bar{R})(Q_{i,t}-\bar{Q}_t)
}{
    \sqrt{\sum_{i=1}^{N}(R_i-\bar{R})^2}
    \sqrt{\sum_{i=1}^{N}(Q_{i,t}-\bar{Q}_t)^2}
},
\]
where $R_i=\operatorname{rank}(s_i)$ and
$Q_{i,t}=\operatorname{rank}(y_{i,t})$. A higher $\rho_t$ indicates better
preservation of the downstream performance ranking.

\noindent\textbf{Pearson correlation.}
We additionally report Pearson correlation to measure linear agreement between
the predicted scores and downstream performance:
\[
r_t =
\frac{
    \sum_{i=1}^{N}(s_i-\bar{s})(y_{i,t}-\bar{y}_t)
}{
    \sqrt{\sum_{i=1}^{N}(s_i-\bar{s})^2}
    \sqrt{\sum_{i=1}^{N}(y_{i,t}-\bar{y}_t)^2}
}.
\]

For both metrics, we report the mean correlation across all $T$ downstream
benchmarks and the correlation with the aggregate downstream performance
$\mathrm{Avg}_i=\frac{1}{T}\sum_{t=1}^{T}y_{i,t}$, where $T$ is the number of
downstream benchmarks. For methods whose native outputs are losses, we negate their scores so that a
larger value consistently indicates a stronger vision encoder.

\subsection{Computational Cost}
\label{sec:computational_cost}
We quantify computational cost using the total number of floating-point operations (FLOPs), providing a hardware-independent measure of evaluation efficiency. For each method, we compute the FLOPs required to evaluate every vision-encoder--LLM pair and report the average across all 210 pairs. The estimate covers the complete evaluation pipeline, including visual and textual feature extraction, feature post-processing, and score computation, as well as probe training and inference when applicable. Averaging over all pairs accounts for differences in encoder architecture, input resolution, and language backbone, yielding a comparable cost estimate for each method.

\section{Further Analysis of Global Geometric Compatibility}
\label{sec:gw_analysis}

We further examine two limitations of GW-based compatibility measures: their
sensitivity to encoder-specific feature geometry and their ability to rematch
samples without preserving observed image--text correspondences.

\subsection{Sensitivity to Encoder Heterogeneity}
\label{sec:gw_heterogeneity}

\begin{figure}[t]
    \centering

    \par\medskip

    \begin{subfigure}[t]{0.48\linewidth}
        \centering
        \includegraphics[width=\linewidth]
        {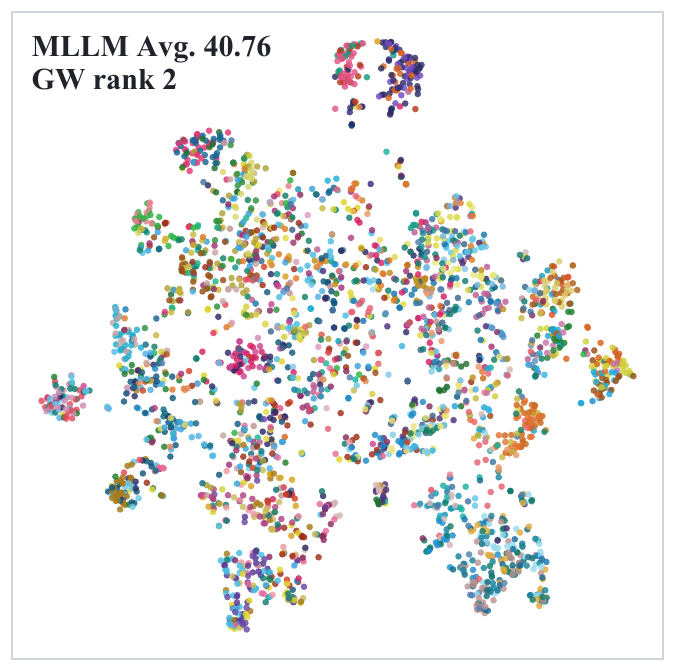}
        \caption{TokLIP feature geometry.}
    \end{subfigure}
    \hfill
    \begin{subfigure}[t]{0.48\linewidth}
        \centering
        \includegraphics[width=\linewidth]
        {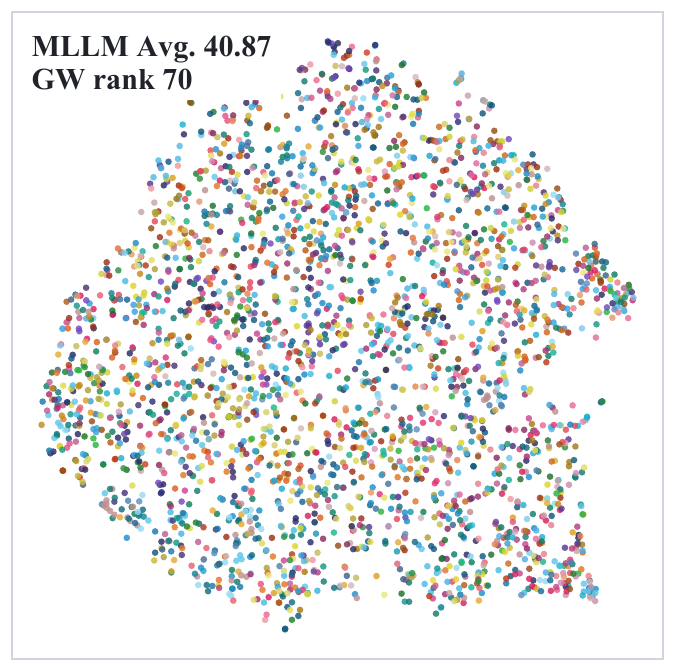}
        \caption{UniAR feature geometry.}
    \end{subfigure}

    \caption{\textbf{Sensitivity of GW to encoder heterogeneity.} Encoders with similar downstream performance can exhibit markedly different feature distributions.}
    \label{fig:gw_heterogeneity}
\end{figure}

\citet{li2026rethinking} use Gromov--Wasserstein (GW) distance to compare the
global geometry of visual and textual representations. However, as shown in
Figure~1, its ranking agreement depends strongly on the candidate pool. GW
performs well in a small, homogeneous pool but becomes less reliable as
heterogeneous encoder families are introduced.
Figure~\ref{fig:gw_heterogeneity}(a) and (b) illustrate this issue: two encoders
with similar downstream performance under
Qwen2.5-1.5B~\citep{yang2024qwen25} exhibit markedly different feature
distributions in t-SNE visualizations~\citep{vandermaaten2008tsne}. Thus,
global geometry may reflect encoder-specific training objectives and output
interfaces beyond downstream MLLM utility.

\subsection{Does Global Geometry Preserve Image--Text Correspondences?}
\label{sec:gw_pairing_details}

A second issue arises from the free coupling optimized by GW. A low discrepancy
may reflect an alternative geometric matching rather than consistency with the
observed image--caption pairs. We examine this distinction by comparing the
optimized GW coupling with a fixed pair-preserving coupling.

\begin{figure}[t]
    \centering
    \begin{subfigure}[t]{0.27\linewidth}
        \centering
        \includegraphics[width=\linewidth]
        {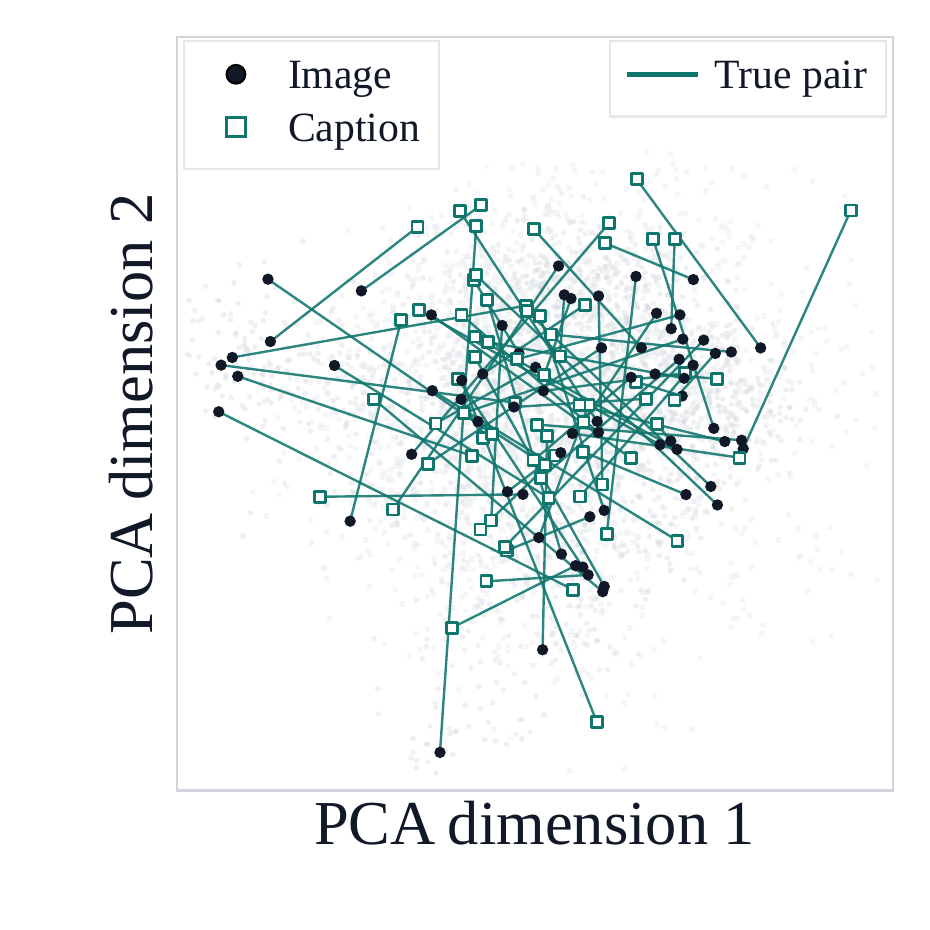}
        \caption{Observed pairing.}
    \end{subfigure}
    \hspace{0.035\linewidth}
    \begin{subfigure}[t]{0.27\linewidth}
        \centering
        \includegraphics[width=\linewidth]
        {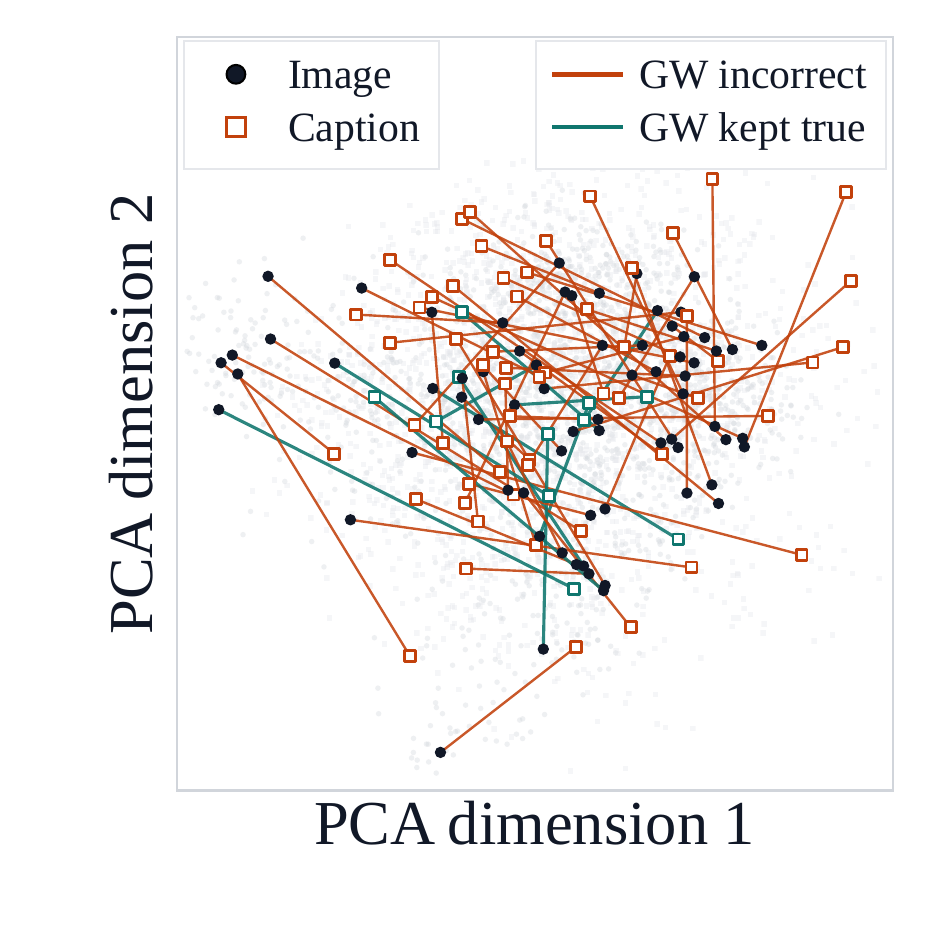}
        \caption{GW-induced rematching.}
    \end{subfigure}
    \hspace{0.035\linewidth}
    \begin{subfigure}[t]{0.27\linewidth}
        \centering
        \includegraphics[width=\linewidth]
        {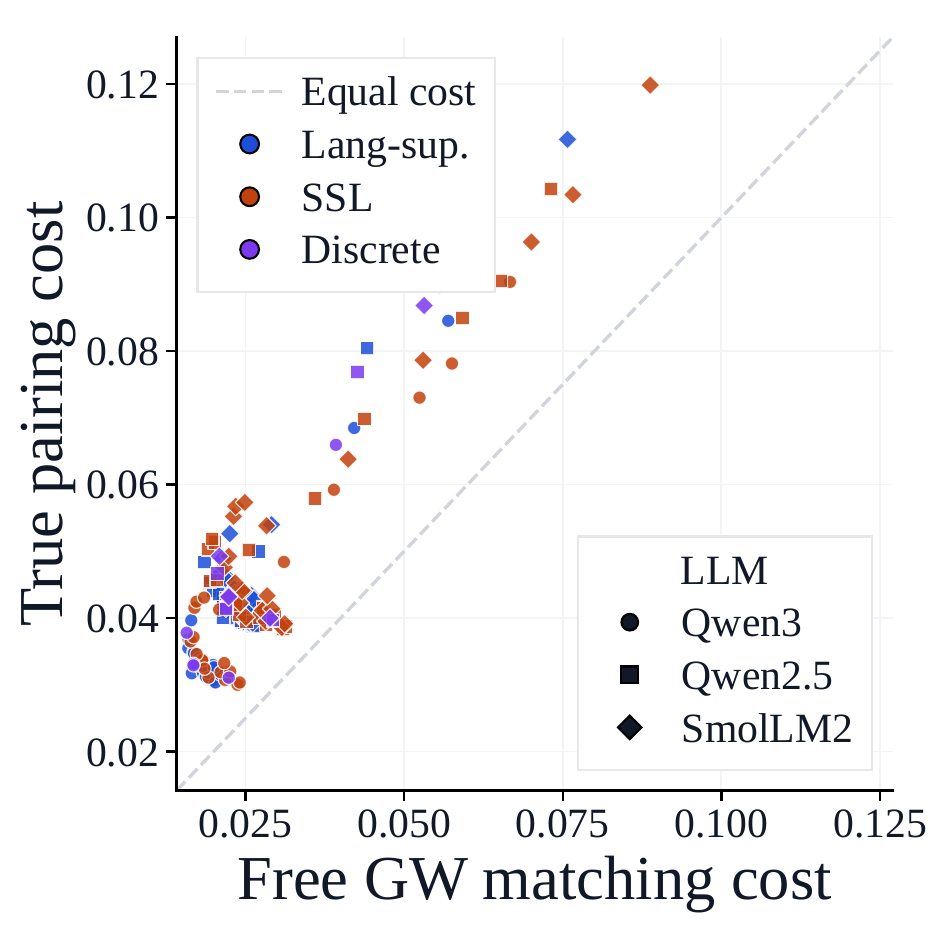}
        \caption{Cost of pair preservation.}
    \end{subfigure}
    \caption{\textbf{Free GW matching can bypass observed cross-modal
    correspondences.}
    (a--b) Observed pairs and maximum-mass GW assignments for a representative
    encoder--language-backbone combination.
    (c) Pair-preserving versus free-matching costs across all 210 combinations.}
    \label{fig:gw_pairing_analysis}
\end{figure}

\noindent\textbf{Data and representations.}
We randomly sample \(n=2{,}000\) image--caption pairs from the COCO 2017
validation set~\citep{lin2014coco} using seed 42. Images are represented by
global features from each of the 70 candidate vision encoders, while captions
are encoded by Qwen3-1.7B, Qwen2.5-1.5B, and SmolLM2-1.7B. All representations
are \(L_2\)-normalized.

\noindent\textbf{GW computation.}
For each of the \(70\times3=210\) encoder--language-backbone combinations, we
solve the exact \(L_1\)-angular GW objective from the image--image and
text--text distance matrices using uniform marginals and a MutualNN
initialization. This produces a free coupling \(T^\star\), which we compare
with the pair-preserving coupling \(T_{\mathrm{pair}}=I_n/n\). We measure
correspondence recovery as
\[
\operatorname{PairAcc}(T^\star)
=
\frac{1}{n}
\sum_{i=1}^{n}
\mathbf{1}\!\left\{
\operatorname*{arg\,max}_{j}T^\star_{ij}=i
\right\}.
\]

\noindent\textbf{Results and visualization.}
Figures~\ref{fig:gw_pairing_analysis}(a) and
\ref{fig:gw_pairing_analysis}(b) visualize SigLIP2 So400m/14 paired with
Qwen3-1.7B. Image and caption features are projected independently to two
dimensions using PCA, after which the caption coordinates are
Procrustes-aligned to the image coordinates for visualization only. We show
the observed assignments \(i\mapsto i\) and the maximum-mass GW assignments
\(i\mapsto\arg\max_j T^\star_{ij}\).

GW extensively reassigns captions: \(\operatorname{PairAcc}\) averages only
\(0.0077\) across all 210 combinations. Moreover, every combination lies above
the equal-cost diagonal in Figure~\ref{fig:gw_pairing_analysis}(c), with the
pair-preserving cost averaging \(1.74\times\) the optimized free-matching cost.
These results show that a low GW discrepancy can arise from sample rematching
rather than preservation of the observed cross-modal correspondences.

Together, these analyses show that GW can be affected by encoder-specific
feature geometry while obtaining low costs through correspondence-agnostic
rematching, limiting its reliability for heterogeneous encoder selection.

\section{Supervised Reference Method}
\label{app:supervised_method}
\ravel targets training-free encoder selection, assuming no downstream performance labels for the candidate encoders. As a complementary analysis, we also consider a supervised setting in which the downstream MLLM performance of a small subset of encoder--language-backbone pairs is available. This setting examines whether previously evaluated candidates can provide additional supervision for selecting among unseen candidates and provides a supervised reference for interpreting RAVEL's results.

Directly learning from frozen encoder features is difficult because the candidate feature spaces are incompatible: their dimensions range from $384$ to $4608$, and their coordinates arise from heterogeneous architectures and pretraining objectives. We therefore develop \textbf{CKA-X}, which represents each candidate using coordinate-agnostic geometric signals. The first measures its centered-kernel-alignment similarity~\citep{kornblith2019cka} to a shared set of reference encoders, while the second measures its cross-modal compatibility with the frozen target language model. A gradient-boosted decision-tree ensemble maps these signals to predicted downstream performance. CKA-X requires downstream labels from the reference candidates but does not require training MLLMs for the held-out candidates.

\subsection{Problem Setup}
\label{sec:supervised_setup}

We now consider the supervised counterpart of \S\ref{sec:method}, keeping its
notation: $\mathcal{D} = \{(x_i, t_i)\}_{i=1}^{n}$ is the paired calibration
data, $E_m$ is a candidate vision encoder, and $L$ is the target language
model. The candidate pool $\mathcal{V} = \{E_m\}_{m=1}^{M}$ consists of the
$70$ vision encoders evaluated in our experiments, each paired with one of the
three language backbones; each $E_m$ now also carries a ground-truth score
$y_m \in \mathbb{R}$, the unweighted mean over the eleven downstream benchmarks
of the MLLM built from $E_m$ with backbone $L$. Let
$Z_m \in \mathbb{R}^{n \times d_m}$ stack the visual representations that
$E_m$ produces on the calibration images (the $Z^{v,m}_i$ of
\S\ref{sec:method}), and let $Z^{t} \in \mathbb{R}^{n \times d_g}$ stack the
text representations that $L$ produces on their paired texts. We evaluate this
by holding each candidate out in turn: for a held-out candidate we predict its
score from its frozen features and the labeled reference pool of the remaining
candidates, $\mathcal{R} = \mathcal{V} \setminus \{\text{held-out}\}$, without
training an MLLM for any candidate. No downstream score of a held-out encoder
is ever used, and the reference pool contains only candidates whose scores are
known.

Raw coordinates cannot support this prediction: padding equalizes only the
dimensionality, not the coordinate systems, so raw features from different
encoders remain numerically incomparable even after zero-padding to a common
width. We therefore represent each candidate by two frame-invariant signals, a
pairwise CKA similarity profile and a cross-modal alignment, and map them to
the score with gradient-boosted decision trees.

\subsection{Pairwise CKA Similarity}
\label{sec:cka_kernel}

To construct a common descriptor for encoders with heterogeneous feature spaces, we compare each candidate with a shared pool of reference encoders using linear centered kernel alignment (CKA)~\citep{kornblith2019cka}. Let
\[
H = I_n - \frac{1}{n}\mathbf{1}\mathbf{1}^\top
\]
be the centering matrix. For candidate \(E_m\), we first \(L_2\)-normalize the rows of \(Z_m\) and construct its centered Gram matrix
\[
G_m = HZ_mZ_m^\top H.
\]
For every candidate pair \((m,m')\), their linear CKA similarity is
\begin{equation*}
  \mathrm{CKA}(m,m')
  =
  \frac{\langle G_m,G_{m'}\rangle_F}
       {\lVert G_m\rVert_F\lVert G_{m'}\rVert_F}
  =
  \frac{\operatorname{tr}(G_mG_{m'})}
       {\sqrt{\operatorname{tr}(G_m^2)\operatorname{tr}(G_{m'}^2)}}
  \in[0,1].
\end{equation*}

Given a reference pool \(\mathcal{R}\), we represent each candidate by its CKA similarities to the reference encoders:
\begin{equation*}
  \mathbf{k}_m
  =
  \bigl(\mathrm{CKA}(m,m')\bigr)_{m'\in\mathcal{R}}.
\end{equation*}
The same reference pool is used for both the candidates fitted by the predictor and the held-out candidates, placing all descriptors in the common space \(\mathbb{R}^{|\mathcal{R}|}\).

CKA is well suited to this setting because it compares the geometry that two encoders induce over the shared images rather than their raw feature coordinates. In particular, it measures whether the encoders organize the images similarly and is invariant to orthogonal transformations, isotropic rescaling, and the addition of zero-valued feature dimensions. It can therefore compare representations with different dimensionalities and coordinate systems without learning an explicit mapping between them.

Moreover, \(\mathbf{k}_m\) is relational: it describes how a candidate is positioned relative to the scored reference encoders. Unlike a single intrinsic representation statistic, this descriptor retains cross-encoder information that allows a predictor fitted on evaluated candidates to transfer to an unseen candidate. We use linear CKA because it is dimension-agnostic, requires no rank-selection hyperparameter, and can be computed directly from the centered Gram matrices.

\subsection{Cross-Modal Alignment}
\label{sec:cross_modal}

The CKA profile characterizes each candidate relative to the reference encoder pool but does not account for the target language model. We therefore augment it with two signals that measure the compatibility between the visual representation and the text representation produced by the frozen target LLM \(L\). Let \(Z^t\in\mathbb{R}^{n\times d_g}\) denote the text representations of the paired captions, obtained by mean-pooling the penultimate-layer hidden states of \(L\).

For each candidate \(E_m\), we compute two scalar statistics. The first is the held-out \(R^2\) of a ridge regression that predicts \(Z^t\) from \(Z_m\). The regression is fitted on one randomly selected half of the calibration pairs and evaluated on the other, using split seed \(42\); its regularization strength is selected by leave-one-out cross-validation within the fitting split. The second statistic is the linear CKA between \(Z_m\) and \(Z^t\), which measures the agreement between their induced sample geometries. Together, these statistics form the cross-modal descriptor
\begin{equation*}
    \mathbf{x}_m
    =
    \left[
        R_m^2,\,
        \mathrm{CKA}(Z_m,Z^t)
    \right]^\top
    \in\mathbb{R}^2.
\end{equation*}

\subsection{Gradient-Boosted Regression}
\label{sec:regression}

We concatenate the relational CKA profile and cross-modal descriptor into
\begin{equation*}
    \mathbf{d}_m
    =
    [\mathbf{k}_m;\mathbf{x}_m]
    \in\mathbb{R}^{|\mathcal{R}|+2}
\end{equation*}
and use gradient-boosted regression trees to predict the downstream MLLM score:
\begin{equation*}
    \hat{y}_m
    =
    b+\eta\sum_{t=1}^{T}h_t(\mathbf{d}_m),
\end{equation*}
where \(b\) is the initial prediction, \(h_t\) is a depth-limited regression tree, \(\eta\) is the learning rate, and \(T\) is the number of boosting rounds. Gradient boosting is well suited to this compact descriptor because it can capture nonlinear threshold effects and interactions between similarity to the reference encoders and compatibility with the target LLM.

We compare gradient boosting with several common alternatives, including ridge regression, a multilayer perceptron, and kernel ridge regression using the CKA matrix as its kernel. All preprocessing and model selection are performed using only the training folds. Gradient boosting achieves the strongest performance for all three language backbones, whereas the multilayer perceptron performs substantially worse. We therefore adopt gradient boosting as the default predictor. Its hyperparameters are fixed across all folds and language backbones: depth-\(2\) trees, \(T=50\) boosting rounds, learning rate \(\eta=0.05\), and subsampling ratio \(0.8\). The resulting model also provides feature-importance estimates for analyzing the contributions of the relational and cross-modal signals.

Scoring a held-out candidate requires only its frozen representations and the downstream scores of the reference pool. Thus, the supervision cost is confined to the reference candidates, and no MLLM needs to be trained for the held-out candidate.

\begin{figure}[t]
  \centering
\includegraphics[width=0.5\linewidth]{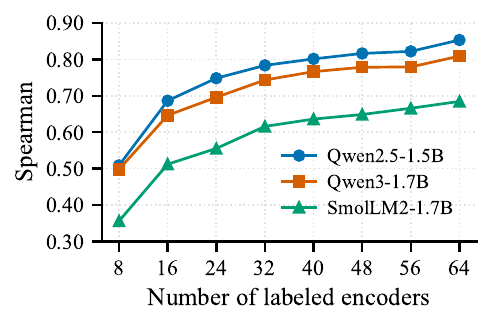}
  \caption{Data efficiency of CKA-X with the gradient-boosted predictor on the
  $70$-encoder pool. Each curve is one target language backbone $L$; the
  $x$-axis is the number of labeled encoders in the reference pool
  $\mathcal{R}$, with the remaining $70-|\mathcal{R}|$ candidates held out.
  Points are pooled Spearman correlations over the held-out predictions of
  $100$ random labeled subsets. }
  \label{fig:labels_with_spearman}
\end{figure}

This cost is set by how many candidates are labeled rather than by which
predictor is used, so we also ask how many labeled encoders CKA-X actually
needs. For each budget
$|\mathcal{R}|\in\{8,16,24,32,40,48,56,64\}$ we draw $100$ random labeled
subsets $\mathcal{R}\subset\mathcal{V}$ (seed $0$), fit the predictor using the
labels of $\mathcal{R}$ alone, and score the remaining candidates
$\mathcal{V}\setminus\mathcal{R}$; the leave-one-out protocol above is the
maximal-budget case $|\mathcal{R}|=M-1=69$. Figure~\ref{fig:labels_with_spearman}
reports the resulting correlation on the held-out candidates. Supervision
transfers quickly across the pool: eight labeled encoders already yield pooled
Spearman correlations of $0.50$, $0.51$ and $0.36$ for Qwen3-1.7B, Qwen2.5-1.5B
and SmolLM2-1.7B, and the curves rise monotonically to $0.81$, $0.85$ and $0.69$
at the largest budget, flattening from about half the pool onward.

\section{Representation post-processing variants}
\label{app:postprocessing-methods}
\noindent\textbf{Protocol.}
The post-processing comparison isolates how feature statistics affect cross-modal neighborhood agreement.  In this diagnostic, each visual representation is obtained by averaging all valid spatial patch tokens of an image before post-processing, while each textual representation is the penultimate-layer last-token feature.  We use the same $n=1{,}000$ LCS-558K examples (seed 42), 70 visual tokenizers, and three language backbones for every variant.  The transformation is fitted independently for each visual tokenizer and each language backbone, so no projection parameters are shared across modalities or models.  After transformation, every sample is row-wise $\ell_2$-normalized, and Top-100 cosine neighborhoods are used to compute the binary MutualNN overlap.  No patch-level matching, rank weighting, hub correction, or chance adjustment is applied in this comparison; only the representation post-processing changes.

\noindent\textbf{Raw.}
This baseline leaves the representation coordinates unchanged and applies only row normalization,
\begin{equation}
    \widetilde{X}_{\mathrm{raw}}=\mathcal{R}(X).
\end{equation}
It therefore preserves the original mean offset, coordinate scales, correlations, and anisotropy, while removing differences in the overall norm of individual samples.

\noindent\textbf{Centered.}
We subtract the feature-wise mean estimated from the fixed evaluation sample and then normalize each row,
\begin{equation}
    \widetilde{X}_{\mathrm{center}}=\mathcal{R}(X-\mathbf{1}\mu^\top).
\end{equation}
Centering removes a common offset shared by the examples but does not alter the relative scale or correlation of different feature dimensions.

\noindent\textbf{Standardized.}
We additionally divide each centered coordinate by its population standard deviation,
\begin{equation}
    \widetilde{X}_{\mathrm{std}}
    =\mathcal{R}\!\left(\frac{X-\mathbf{1}\mu^\top}{\sigma+10^{-5}}\right),
\end{equation}
where the division is coordinate-wise.  This gives the original coordinates comparable marginal scales and reduces the influence of unusually high-variance dimensions, but it does not decorrelate them and remains tied to the encoder's native coordinate basis.

\noindent\textbf{PCA only.}
We project the centered features onto the retained principal subspace and normalize the resulting scores,
\begin{equation}
    \widetilde{X}_{\mathrm{PCA}}=\mathcal{R}(Z).
\end{equation}
This rotation decorrelates the retained coordinates and truncates the low-variance tail beyond rank $r$, but it keeps the variance ordering within the retained subspace.  Consequently, leading principal components can still dominate cosine neighborhoods.

\noindent\textbf{ABTT.}
All-but-the-Top (ABTT)~\citep{mu2017all} starts from the same PCA scores but discards the first $q$ highest-variance directions before row normalization.  With the fixed setting $q=4$ used in the figure,
\begin{equation}
    \widetilde{X}_{\mathrm{ABTT}}
    =\mathcal{R}\!\left(Z_{[:,q+1:r]}\right),\qquad q=4.
\end{equation}
ABTT directly removes a small number of dominant directions that may encode broad, non-specific variation.  Unlike whitening, however, it deletes those directions entirely, so its behavior depends on the chosen removal count $q$ and can deteriorate when informative principal components are removed.

\noindent\textbf{PCA whitening.}
We retain the same PCA subspace as PCA only, but rescale each component by its regularized standard deviation,
\begin{equation}
    \widetilde{X}_{\mathrm{white}}
    =\mathcal{R}\!\left(
        Z\,\operatorname{diag}\big((\lambda_1+10^{-4})^{-1/2},\ldots,
        (\lambda_r+10^{-4})^{-1/2}\big)
    \right).
\end{equation}
Whitening therefore equalizes the variances of the retained principal directions rather than removing the leading ones.  The additive $10^{-4}$ term limits the amplification of small-eigenvalue directions, and the rank cap removes the least reliable low-variance tail before rescaling.

\begin{figure}[t]
    \centering
    \includegraphics[width=0.5\linewidth]
    {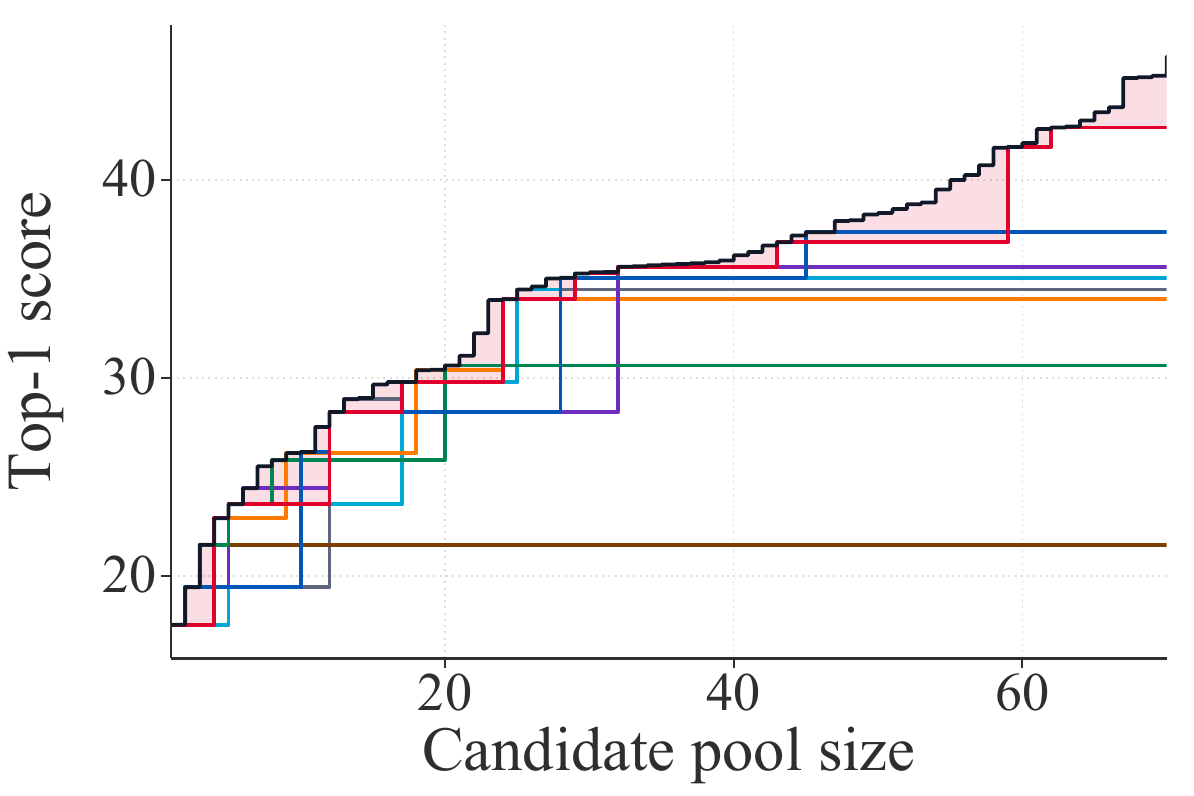}
    \caption{\textbf{Selection robustness across candidate-pool sizes.}
    Ground-truth performance of the encoder selected by each method as
    increasingly strong candidates enter the pool.}
    \label{fig:robustness_candidate_pool}
\end{figure}

\subsection{Baseline Methods}
\label{app:baseline}
\begin{itemize}
    \item Linear probing~\citep{caron2021dino} trains a linear classifier on frozen visual representations and uses image-classification accuracy as the encoder score.
    \item $k$-nearest-neighbor classification (KNN; \citealp{wu2018unsupervised}) predicts image labels from their nearest neighbors in the visual representation space and uses classification accuracy as the encoder score.
\end{itemize}

We next consider cross-modal metrics that compare visual and textual representation spaces without training the corresponding MLLM:
\begin{itemize}
    \item Representational similarity analysis (RSA; \citealp{kriegeskorte2008rsa}) measures the correlation between the pairwise dissimilarity structures of the visual and textual representations, while canonical correlation analysis (CCA; \citealp{hotelling1936cca}) learns linear projections of the two spaces that maximize their correlation.
    \item Gromov--Wasserstein (GW; \citealp{li2026rethinking}) compares visual and textual representations by matching their within-modality distance structures through optimal transport, while mutual \(k\)-nearest-neighbor alignment (MutualNN; \citealp{huh2024platonic}) measures the overlap between nearest-neighbor sets
constructed in the visual and textual representation spaces.
    
\end{itemize}

Finally, we include several methods that require additional model training or fitting. Because of their substantially higher computational or supervision requirements, these methods may be less suitable as practical evaluation metrics for large-scale vision encoder selection. Nonetheless, we include them to provide stronger reference points:
\begin{itemize}
    \item Mid-training loss performs the first-stage mid-training for each encoder--language-model pair using the LLaVA training recipe and uses the loss at the final optimization step as an estimate of downstream performance.
    \item Alignment probing~\citep{zhang2025alignment} uses labeled image--text data to learn linear mappings between the visual and textual spaces and evaluates their alignment through cross-modal retrieval.
    \item AC Policy~\citep{yang2025law} fits a polynomial regressor that maps alignment and visual correspondence scores to the observed downstream performance of a subset of vision encoders.
\end{itemize}

\section{Robustness under a Selection-Based Metric}
\label{app:selection_robustness}

We evaluate whether each method reliably selects a strong encoder as the
candidate pool changes. For each language backbone, we order the 70 encoders
by ground-truth downstream performance and progressively expand the pool from
the worst encoder to the full set. At each pool size, every method selects its
highest-scoring encoder, whose ground-truth performance is reported alongside
that of the best available encoder (oracle).

As shown in Figure~\ref{fig:robustness_candidate_pool}, \momo{} closely tracks
the rising oracle throughout the expansion, whereas competing methods plateau
as stronger encoders become available. This demonstrates its robustness to
changes in candidate-pool size and composition.